\documentclass[12pt]{elsarticle}
\journal{Journal of Biomedical Informatics}
\usepackage{amsmath,amssymb, amsfonts}
\usepackage{helvet}

\usepackage[left=1in,top=1in,right=1in, bottom=1in]{geometry}
\usepackage{setspace}
\usepackage{todonotes}
\usepackage{makecell}

\newcommand{\sect}{\textbf{Methods}}
\newcommand{\fig}{\textbf{Fig.}}

\newcommand{\suppsec}{\textbf{Supplementary Section}}

\newcommand{\supfig}{\textbf{Supplementary Fig.}}

\newcommand{\mf}{\mathbf}

\newcommand{\model}{MixEHR-SurG}

\usepackage{upgreek}

\usepackage{xr}
\usepackage{url}
\usepackage{multirow}
\usepackage{algpseudocode}
\usepackage{algorithm2e}
\LinesNumbered
\RestyleAlgo{algoruled} 

\usepackage[T1]{fontenc}    
\usepackage{url}          
\usepackage{nicefrac}      
\usepackage{microtype}    
\usepackage{caption}
\usepackage{subcaption}

\usepackage{lineno}
\usepackage{longtable}
\usepackage{lscape}
\usepackage[final]{pdfpages}
\definecolor{DarkRed}{rgb}{0,0,1}
\newcommand{\hi}{\textcolor{black}}
\newcommand{\hihi}{\textcolor{black}}

\usepackage{adjustbox}
\usepackage{bm}

\usepackage{multirow}
\usepackage{longtable}
\usepackage{colortbl}

\usepackage{lineno,hyperref}

\begin{document}

\begin{frontmatter}

\title{\model: a joint proportional hazard and guided topic model for inferring mortality-associated topics from electronic health records}
\author[1,2]{Yixuan Li}
\author[1,2,4]{Archer Y. Yang\corref{coraut}}
\author[3]{Ariane Marelli\corref{coraut}} 
\author[2,4]{Yue Li\corref{coraut}}

\address[1]{Department of Mathematics and Statistics, McGill University, Montreal, Canada}
\address[2]{Mila - Quebec AI institute, Montreal, Canada}
\address[3]{McGill Adult Unit for Congenital Heart Disease (MAUDE Unit), McGill University of Health Centre, Montreal, Canada}
\address[4]{School of Computer Science, McGill University, Montreal, Canada}
\cortext[coraut]{Co-corresponding: 
archer.yang@mcgill.ca;
ariane.marelli@mcgill.ca;  yueli@cs.mcgill.ca}
\date{}





\begin{abstract}
\hi{Survival models can help medical practitioners to evaluate the prognostic importance of clinical variables to patient outcomes such as mortality or hospital readmission and subsequently design personalized treatment regimes. Electronic Health Records (EHRs) hold the promise for large-scale survival analysis based on systematically recorded clinical features for each patient. However, existing survival models either do not scale to high dimensional and multi-modal EHR data or are difficult to interpret.}
\hi{In this study, we present} a supervised topic model called \model~to simultaneously integrate heterogeneous EHR data and model survival hazard.
Our contributions are three-folds: (1) integrating EHR topic inference with Cox proportional hazards likelihood; (2) integrating patient-specific topic hyperparameters using the PheCode concepts such that each topic can be identified with exactly one PheCode-associated phenotype; (3) multi-modal survival topic inference. This leads to a highly interpretable survival  topic model that can infer PheCode-specific phenotype topics associated with patient mortality. We evaluated \model~using a simulated dataset and two real-world EHR datasets: the Quebec Congenital Heart Disease (CHD) data consisting of 8,211 subjects with 75,187 outpatient claim records of 1,767 unique ICD codes; the MIMIC-III consisting of 1,458 subjects with multi-modal EHR records.
Compared to the baselines, \model~achieved a superior dynamic AUROC for mortality prediction, with a mean AUROC score of 0.89 in the simulation dataset and a mean AUROC of 0.645 on the CHD dataset. Qualitatively, \model~associates severe cardiac conditions with high mortality risk among the CHD patients after the first heart failure hospitalization and critical brain injuries with increased mortality among the MIMIC-III patients after their ICU discharge.
Together, the integration of the Cox proportional hazards model and EHR topic inference in \model~not only leads to competitive mortality prediction but also meaningful phenotype topics for in-depth survival analysis. 
The software is available at GitHub: \url{https://github.com/li-lab-mcgill/MixEHR-SurG}.
\end{abstract}


\end{frontmatter}




\section{Introduction}
\label{background and significance}


The rapid adoption of Electronic Health Records (EHRs) \cite{jiang2023pre} enables systematic investigation of phenotypes and their comorbidity \cite{smoller2018use, alzoubi2019review, shivade2014review, jensen2012mining}. EHR include rich phenotypic observations of patient subjects from physician and nursing notes to diagnostic codes and prescription. 
One important application of EHR is to \textit{detect} and \textit{understand} the risk of adverse events such as death based on the recent health history of the patient \cite{jensen2017analysis}. Accurate detection will enable efficient resource allocation for the high-risk patients and can cost-effectively  
save many lives \cite{javaid2022significance}. Understanding the mortality risk is equally important as it can inform practitioners for subsequent intervention. 
Many machine learning methods were developed recently for predicting adverse events such as mortality and unplanned emergency re-admission \cite{miotto2016deep, ranganath2016deep, lee2018deephit, shin2021machine}. However, the progress on this front has been hindered by the lack of an interpretable approach that can distill interpretable phenotypic concepts relevant to the outcome of interest while having competitive detection precision on those events.

Predicting mortality events using EHRs has been a long-standing challenge due to the large search space of causal events. Survival analysis models have evolved beyond traditional Cox proportional hazards (PH) models \cite{cox1972regression} to include sophisticated techniques capable of handling complex, high-dimensional data. For instance, the kernel Cox regression method \cite{li2002kernel} extends the Cox model by incorporating kernel methods, allowing for a nuanced understanding of patient survival in relation to a broader range of clinical factors. Random Survival Forests \cite{ishwaran2008random} and LASSO-penalized Cox models \cite{tibshirani1997lasso} were developed for high-dimensional data, enhancing the predictive accuracy and interpretability of survival outcomes. These advancements represent significant strides in survival analysis, enabling more precise and comprehensive evaluations of patient data. While these have set the foundational benchmark, they sometimes sidestep the complex, patient-specific nuances. Recently, deep learning methods like DeepSurv \cite{ranganath2016deep} and neural multi-task logistic regression \cite{miotto2016deep} have entered the fray, harnessing the power of neural networks to predict patient survival with greater accuracy. However, these methods are hard to interpret and often require external approaches to explain their prediction \cite{lundberg2017unified,chen2022explaining,lundberg2020local}.

Topic models are a family of Bayesian models \cite{blei2003latent}. In our context, we treat patients as documents and their EHR codes as tokens. Topic models infer the topic mixture of each document, the latent topic for each token, and a set of latent topic distributions. Here the topic mixture represents the mixture of phenotype of the patient and the set of topic distributions represent the set of phenotypic distributions over the EHR codes. Despite the simple generative process, topic models are effective in distilling phenotype concepts from the EHR data \cite{li2020inferring,song2021supervised,song2022automatic}. Recently, we developed a guided topic model called Mixture of EHR Guided (MixEHR-G) \cite{ahuja2022mixehr}, which specifies the topic hyperparameters based on the high-level phenotype codes (i.e., PheCodes) observed in the patients. As a result, each topic is identifiable with known phenotype codes, thereby improving the down-stream analysis. However, MixEHR-G does not have the ability to predict mortality. \hi{To address this challenge, we aim to develop a model that leverages EHR data for two primary purposes: (1) inferring mortality risk of patients from their multi-modal EHRs; (2) identifying phenotype concepts of disease comorbidity in order to explain high risk of mortality.}

\begin{figure}[t!]
\includegraphics[width=1.0\textwidth]{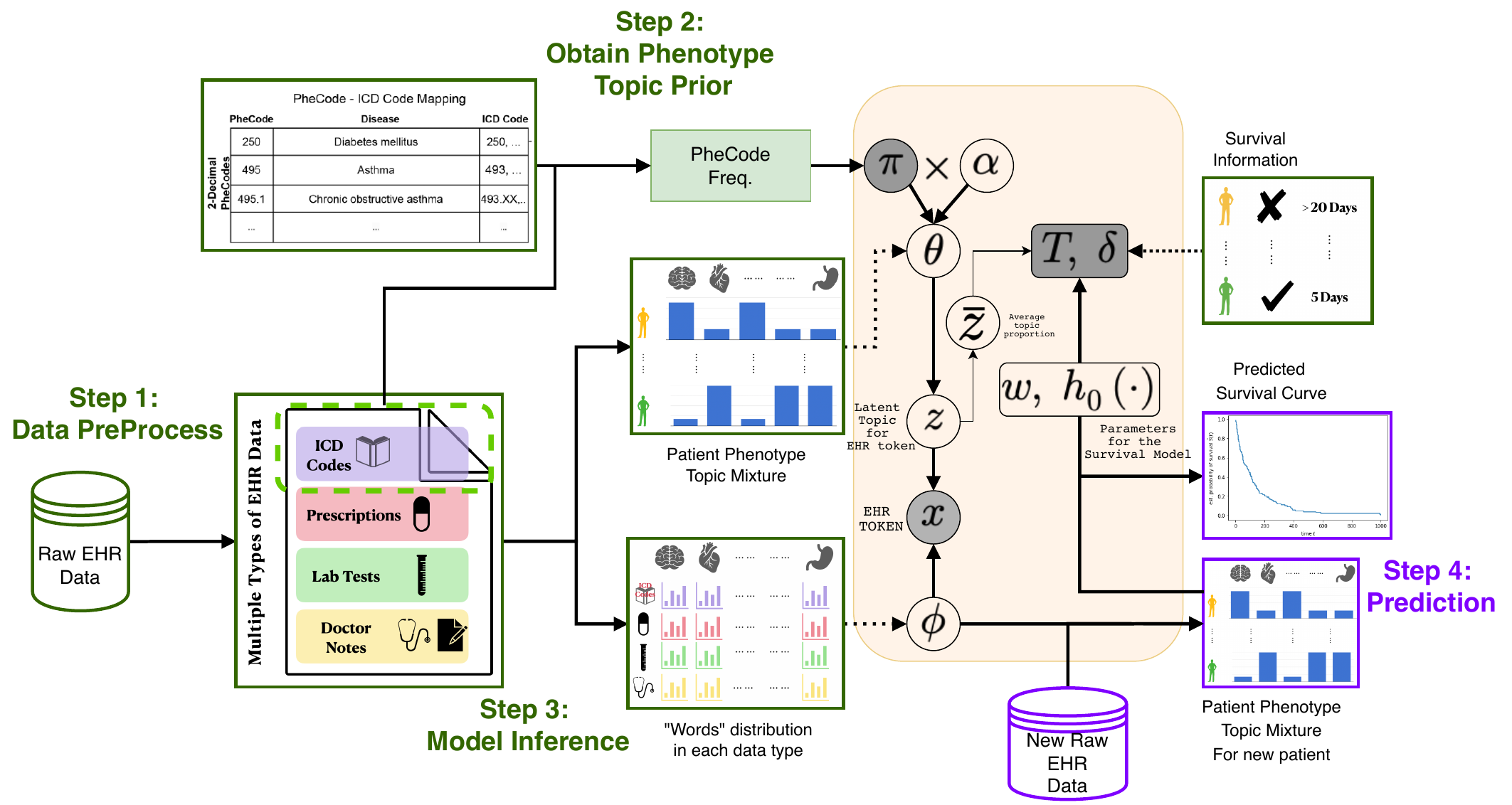}
\caption{\model~overview. \model~consists of four main steps. The training process is highlighted in green, and the prediction process is depicted in purple. In Step 1, we prepossess and aggregate raw EHR data for each patient $j$.  Step 2 involves determining a $K$-dimensional phenotype topic prior, $\boldsymbol{\uppi}_j = (\pi_{j1}, \ldots, \pi_{jK})$, for each patient. Step 3 infers phenotype topic distribution $\boldsymbol{\upphi}_k^{(m)} \in \mathbb{R}^{V^{(m)}}$ for EHR type $m$ in topic $k$ (i.e., the model parameters of \model). This requires inferring the latent topic assignment $z_{ji}\in\{1,\ldots,K\}$ for each EHR token $i$ in patient $j$. In Step 4, the trained model is applied to predict personalized survival function for new patient. The details of the probabilistic graphical model is depicted in \fig~\ref{fig:MixEHRs_diagrams}.}\label{fig:MixEHR_SurG}
\end{figure}

In this study, we present MixEHR Survival Guided (\model; \fig~\ref{fig:MixEHR_SurG}). \model~is an extension of MixEHR \cite{li2020inferring} and MixEHR-G  \cite{ahuja2022mixehr} and designed to integrate survival information and high-dimensional EHRs data via a supervised topic model framework \cite{dawson2012survival}. 
\hi{Our contributions are three-folds: (1) integrating EHR topic inference with Cox proportional hazards likelihood; (2) integrating patient-specific topic hyperparameters using the PheCode concepts such that each topic can be identified with exactly one PheCode-associated phenotype; (3) multi-modal survival topic inference.}
As a result, \model~can perform guided phenotype topic inference and survival risk analysis simultaneously. 
We perform comprehensive evaluations of \model, benchmarking on both its predictive accuracy for patient survival times and its ability to generate meaningful survival-related phenotype topics. In our simulation study, \model~not only accurately predicts survival times but also identifies true survival topics. When applied to the real-world Quebec Congenital Heart Disease (CHD) dataset and the MIMIC-III ICU dataset, \model~excels in predicting survival times and produces meaningful mortality-related phenotype topics. In the CHD dataset, \model~reveals cardiac-related phenotypes as significant mortality risk factors after the first onset of heart failure. In the MIMIC-III dataset, \model~identifies critical neurological conditions as one of the key mortality indicators. 




\begin{figure}[t!]
\centering\includegraphics[width=1.0\textwidth]{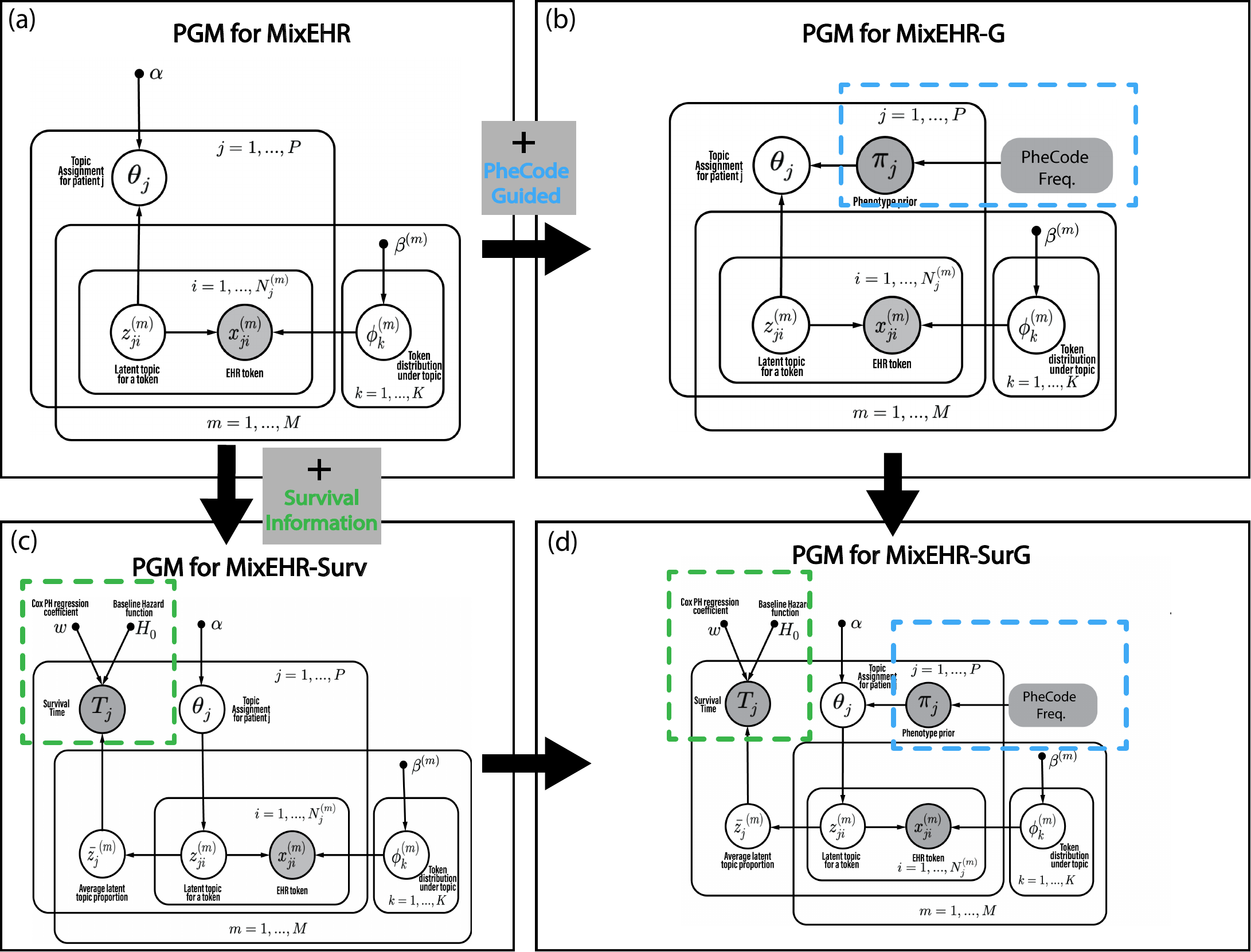}
\caption{Probabilistic graphical model (PGM) illustration of four models variants. 
(a) PGM for MixEHR. We first generate topic distributions $\boldsymbol{\upphi}_k^{(m)}$ for each topic $k$ and document type $m$, then we generate of a $K$-dimensional topic proportion $\boldsymbol{\uptheta}_j$ for every patient $j$. Finally, we generate latent topics $z_{ji}^{(m)}$ and corresponding words $x_{ji}^{(m)}$ for each EHR token. 
(b) PGM for MixEHR-G. We infer patient specif PheCode-Guided topic prior $\boldsymbol{\uppi}_j$ for each patient $j$ and used it as Dirichlet hyperparameters for the patient topic mixture $\boldsymbol{\uptheta}_j$ enclosed by a blue dashed rectangular. 
(c) PGM for MixEHR-Surv. For each patient $j$, we obtained the survival time $T_j$ and employed the Cox proportional hazards (PH) model with coefficient $\mathbf{w}$ and baseline hazard function $h_0(\cdot)$ to guide the learning of topics, as enclosed by a green dashed rectangular. 
(d) PGM for the proposed \model. We combine both PheCode-Guided prior and survival information into one single model. The resulting model can use the guided phenotype topics to model the Cox PH of survival likelihood}\label{fig:MixEHRs_diagrams}
\end{figure}
 
\section{Methods}
\label{methods and materials}




\subsection{MixEHR}\label{MixEHR-model}
This section briefly reviews MixEHR \cite{li2020inferring}.
EHR includes a collection of medical documents of $M$ types, indexed by $m=1, \ldots, M$, such as ICD codes, drug codes, and clinic notes, etc. These documents provide a comprehensive overview of patients' clinical histories and examination results, which reflects personal health conditions. For document type $m$, a list of EHR features, indexed by $v=1,\ldots V^{\left( m\right) }$, encompasses all potential unique EHR features that are collected for that specific document type present in the dataset. For patient $j\in \left\{ 1,\ldots ,P\right\}$, the EHR document of type $m$ contains $N^{\left( m\right) }_{j}$ tokens, and each token is represented as $x_{ji}^{\left( m\right) }$, for $i=1,\ldots, N^{\left( m\right) }_{j}$. 
In the context of topic modeling, the feature distribution of document type $m$ under topic $k$ is denoted as $\boldsymbol{\upphi}_{k}^{\left( m\right) } =[\phi_{kv}^{\left( m\right) } ]_{v=1}^{V^{\left( m\right) }}\in \mathbb{R}^{V^{(m)}}$. These weights are derived from a Dirichlet distribution, with an unknown hyperparameter $\boldsymbol{\upbeta}^{(m)}\in \mathbb{R}^{V^{(m)}}$. Additionally, the model assumes a specific topic assignment, represented as $\boldsymbol{\uptheta}_j\in \mathbb{R}^{K}$, for each patient $j$, which is also derived from a Dirichlet distribution, with a $K$-dimensional hyperparameter $\boldsymbol{\alpha}$. For every EHR token $x_{ji}^{\left( m\right) }$, with a latent topic assignment represented as $z_{ji}^{\left( m\right) }$, MixEHR has the following generative process (\fig~\ref{fig:MixEHRs_diagrams}a): 
\begin{enumerate}
    \item Generate the feature distribution $\boldsymbol{\upphi}_{k}^{\left( m\right)} \sim \text{Dir}\left(\boldsymbol{\upbeta}^{(m)}\right)$ for topic $k=1,\ldots,K$ and type $m=1,\ldots, M$.
    \item For each patient $j=1,\ldots,P$, sample a $K$-dimensional topic mixture: $\boldsymbol{\uptheta}_{j}\sim\text{Dir} \left(\boldsymbol{\alpha}\right)$.
    \begin{enumerate}
        \item For each of the EHR token $x_{ji}^{\left( m\right) }$ for $i = 1, \ldots, N^{\left( m\right) }_{j}$,  $j=1,\ldots,P$ and  $m=1,\ldots, M$:
        \begin{enumerate}
            \item Sample a latent topic for token $i$: $z_{ji}^{\left( m\right) }\sim\text{Mul} \left(\boldsymbol{\uptheta}_{j}\right).$ 
            \item Sample a word for token $i$: $x_{ji}^{\left( m\right) }\sim\text{Mul}\left(\boldsymbol{\upphi}_{z_{ji}^{\left( m\right) }}^{\left( m\right) }\right).$
        \end{enumerate}
    \end{enumerate}
\end{enumerate}

The posterior distributions of $\boldsymbol{\uptheta}_{j}$ and $\boldsymbol{\upphi}_{k}^{\left( m\right) }$ are approximated by  the collapsed mean-field variational inference method   \cite{teh2006collapsed}.
Although MixEHR is useful for multi-modal topic inference, it does not directly predict a target phenotype of interest. \cite{song2021supervised} proposed the MixEHR-S model in their study, which enables supervised topic infernece for predicting a binary phenotype label.  Nevertheless, time-to-event outcomes for survival analysis are crucial in medical research and clinical applications. Consequently, we seek to expand the MixEHR family to survival-supervised disease topic learning.

\subsection{MixEHR-G}\label{MixEHR-G}
The data generative process (\fig~\ref{fig:MixEHRs_diagrams}b) assumes that for each patient $j$, a set of noisy phenotype label are observed based on a phenotype reference such as the Phenotype Code or PheCode \cite{wei2017evaluating}. Let $\mathbf{u}_j \in \{0,1\}^{K}$ be a binary vector of observed phenotype labels in patient $j$. The topic mixture is sampled from a Dirichlet distribution $\boldsymbol{\uptheta}_j \sim \text{Dir}(\boldsymbol{\uppi}_j)$, where $\pi_{jk} \equiv p(y_{jk} = 1 \mid u_{jk})$, where $y_{jk}$ is a binary latent variable indicating presence or absence of phenotype $k$ for patient $j$. We infer the posterior distribution of $y_{jk}$ using two-component univariate mixture models as described in \suppsec~\ref{sec:prior}. 
In a nutshell, topic $k$ with the observed phenotype label support in patient $j$ will have relatively higher mixture proportion of $\theta_{jk}$ than those topics without the phenotype label support. The rest of the data generative process is identical to MixEHR.

\subsection{MixEHR-SurG}\label{MixEHR-SurG-model}
Our objective is to identify phenotype topics that are informative of patient survival time. To this end, we extend MixEHR-G to integrate survival information. Let $Y$ be the survival time for a patient, i.e., the time until a specific event occurs. In many applications, such as clinical studies, the survival time of a patient may not be known exactly. For example, a patient may not experience the event before the study ends or dropout during the study period (i.e., censored). Let $C$ be the censoring time. The actual observed time $T$ is either the survival time or the censoring time, whichever comes first, i.e., $T=\min(Y,C)$.  Let $\delta =\mathbb{I}(Y \leq C) \in\left\{0,1 \right\}$ be the censoring status, where $\delta=1$ indicates that $Y$ is observed and $0$ otherwise.

The survival function $S(t) = P(T > t)$ outputs the probability of survival beyond time $t$: 
    $$ 
        S(t)=\exp \left[-H\left(t\right)\right]
    $$
where $H(t)$ is the cumulative hazard function, defined as $H(t)=\int_0^t h(u) \mathrm{d} u$. This function accumulates the hazard function $h(u)$, over the interval $[0,t]$. The hazard function $h(u)$ typically represents the instantaneous risk of the event (such as failure or death) occurring at time $u$.

Here we use a semi-supervised Cox PH model \cite{cox1972regression} for the hazard function. We further assume that the latent topic assignments can influence the survival response  $T_j$ (i.e., observed survival time for patient $j$). Specifically, we first compute the topic proportion $\bar{\mf{z}}_j$ for each patient $j=1,\ldots,P$: 
    $$
        \bar{\mf{z}}_j=[\bar{z}_{jk}]_{k=1}^{K}=\left[\frac{\sum_{m=1}^{M}\sum_{i=1}^{N_{j}^{(m)}}\mathbb{I}(z_{ji}^{\left( m\right) }=k)}{\sum_{m=1}^{M}N_{j}^{\left( m\right) }}\right]_{k=1}^K
    $$
where $\bar{\mf{z}}_j$ can be viewed as the estimate of $\boldsymbol{\uptheta}_{j}$.

Next, the survival time $T_j$ corresponds to the Cox proportional hazards (PH) model with a system-wide $K$-dimensional Cox PH regression coefficients $\mf{w}$ (fixed but unknown). The baseline hazard function is defined as:
$\ensuremath{h_{0}\left(\cdot\right)}$, for each patient $j=1,\ldots,P$: 
    $$
        h\left(T_{j}|\bar{\mf{z}}_j\right)=h_{0}\left(T_{j}\right)\exp\left\{ \mf{w}^{\top}\bar{\mf{z}}_j\right\} 
    $$
The preceding generative process of \model~is the same as MixEHR-G (\fig~\ref{fig:MixEHRs_diagrams}d). Lastly, as one of the simplified model, we also implemented MixEHR-Surv (\fig~\ref{fig:MixEHRs_diagrams}c), which is a survival-supervised MixEHR without using the PheCode guide.

\subsection{MixEHR-SurG model inference}\label{sec:MixEHR-SurG}
As depicted in \fig~\ref{fig:MixEHR_SurG}, \model~combines MixEHR-G and survival topic model into a single model. The joint-likelihood function is:
\begin{align*}
    p\left(\mf{T}, \boldsymbol{\updelta}, \mathcal{X}, \mathcal{Z}, \boldsymbol{\uptheta}, \boldsymbol{\Phi} \mid \boldsymbol{\upalpha}, \boldsymbol{\uppi}, \mathcal{B}, h_0(\cdot) , \mf{w}\right)
    = 
    p(\boldsymbol{\uppi} \mid \mf{U})
    p(\mathcal{X}, \mathcal{Z}, \boldsymbol{\uptheta}, \boldsymbol{\Phi} \mid \boldsymbol{\upalpha}, \boldsymbol{\uppi}, \mathcal{B})     
    p\left(\mf{T}, \boldsymbol{\updelta} \mid \mathcal{Z}, h_0(\cdot) , \mf{w}\right)
\end{align*}
The first term is the prior term $\boldsymbol{\uppi}$ for the phenotype topic, which we separately infer using 2-component mixture univariate model on the Phecode counts matrix $\mf{U}$ for each PheCode-guided topic as detailed in the \suppsec~\ref{sec:prior}. 
The second term is the unsupervised part of the likelihood and the same as the MixEHR-G \cite{li2020inferring}. 
The third term is the survival supervised component of the model. We use the Cox PH model with elastic net penalization (i.e., L1 + L2 norm) \cite{simon2011regularization} for the survival
coefficients. 

While we fit the first term separately and fix $\boldsymbol{\uppi}$ to the expected value, we jointly fit the second and the third term of the joint likelihood. Specifically, the full likelihood function of the penalized Cox PH model is obtained by incorporating Breslow's estimate of the baseline hazard function and the penalty term with the hyperparameter $\lambda_1$ for L1 norm and $\lambda_2$ for L2 norm.
\begin{align*}
 & p(\mf{T},\boldsymbol{\updelta}\mid\mathcal{Z},h_{0}\left(\cdot\right),\mf{w})\\
= & \prod_{j=1}^{P}\left\{ \left[h_{0}\left(T_{j}\right)\exp\left(\mf{w}^{\top}\bar{\mf{z}}_{j}\right)\right]^{\delta_{j}}\exp\left[-H_{0}\left(T_{j}\right)\exp\left(\mf{w}^{\top}\bar{\mf{z}}_{j}\right)\right]\right\} \exp\left\{ -\lambda_{2}\|\mf{w}\|_{2}^{2}-\lambda_{1}\|\mf{w}\|_{1}\right\} 
\end{align*}
Note that here $\mf{w}$ is not a variable and there is no prior distribution. The second term was added only for regularization purpose using Elastic Net (Eq. \eqref{eq:survival_coef}; \textbf{\ref{sgd}}).

We will first integret out $\boldsymbol{\uptheta}$
and $\boldsymbol{\Phi}$ to achieve more accurate and efficient inference, due to the conjugacy of Dirichlet variables $\boldsymbol{\uptheta}$
and $\boldsymbol{\Phi}$ to the multinomial likelihood variables $\mathcal{X}$
and $\mathcal{Z}$ \cite{teh2006collapsed}. Then, the ELBO for the current marginal distribution for the observed data is:
\begin{align}\label{eq:elbo}
    \mathcal{L}_{ELBO}
    = \mathbb{E}_q\left[\log p\left(\mf{T}, \boldsymbol{\updelta}, \mathcal{X}, \mathcal{Z} \mid \boldsymbol{\alpha}, \boldsymbol{\uppi}, \mathcal{B}, h_{0}\left(\cdot\right)\right), \mf{w}\right]
    - \mathbb{E}_q\left[\log q(\mathcal{Z})\right]
\end{align}
where we assume a mean-field variational distribution for the topic assignments:
\begin{align*}
    q(\mathcal{Z}) = \prod_{m=1}^M\prod_{j=1}^{P}\prod_{i=1}^{N_j^{(m)}} q(z_{ji}^{(m)}) = 
    \prod_{m=1}^M\prod_{j=1}^{P}\prod_{i=1}^{N_j^{(m)}} \prod_{k=1} ^K \left(\gamma^{(m)}_{jik}\right)^{[z_{ji}^{(m)}=k]}
\end{align*}
Maximizing the Evidence Lower Bound (ELBO) with respect to $\gamma_{j i k}^{(m)}$ is equivalent to computing the conditional expectation of the variable $z_{j i}^{(m)}=k$ given the estimates for other tokens. There exists an efficient closed-form update expression (\suppsec~\ref{sgd}):
\begin{align}
\gamma_{jik}^{\left(m\right)} & \propto\exp\left\{ \mathbb{E}_{q\left(z_{(j,-i)}^{\left(m\right)}\right)}\left[\log p\left(T_{j},\delta_{j}\mid z_{(j,-i)}^{\left(m\right)},z_{ji}^{\left(m\right)}=k,{h}_{0}(T_{j}),\mf{w}\right)\right]\right\} \nonumber \\
 & \qquad\times\left(\alpha_{k}\pi_{jk}+\left[n_{j\bullet k}^{(\bullet)}\right]_{(j,-i)}\right)\frac{\beta_{x_{ji}^{(m)}}^{(m)}+\left[n_{\bullet x_{ji}^{(m)}k}^{(m)}\right]_{(-j,-i)}}{\sum_{v}\beta_{v}^{(m)}+\left[n_{\bullet vk}^{(m)}\right]_{(-j,-i)}},\label{eq:gamma}
\end{align} where the subscript $\left(j,-i\right)$ indicates excluding token $i$ of patient $j$ when calculating its own expectation and the coordinate sufficient statistics are:
$$
n_{\bullet vk}^{(m)}=\sum_{j=1}^{P}\sum_{i=1}^{N_{j}^{(m)}}\mathbb{I}\left[x_{ji}^{(m)}=v,z_{ji}^{(m)}=k\right],
$$

$$
n_{j\bullet k}^{(\bullet)}=\sum_{m=1}^{M}\sum_{i=1}^{N_{j}^{(m)}}\mathbb{I}\left[z_{ji}^{(m)}=k\right].
$$

This equation is derived under the principle that the Kullback-Leibler (KL) divergence reaches its minimum when the approximation of the variational parameter matches the expectation under all other known latent variables. Additionally, the hyperparameters $\boldsymbol{\alpha}$'s and $\boldsymbol{\upbeta}^{(m)}$'s updates are refined through empirical Bayes by optimizing ELBO given the variational estimates of the topic assignments $\gamma_{jik}^{\left(m\right)}$'s. Detailed derivation are described in \suppsec~\ref{sgd}.

Given the variational expectation of $\mathcal{Z}$, the Cox regression coefficients $\mf{w}$ are fit via penalized log likelihood of $\log p(\mf{T},\boldsymbol{\updelta} \mid \mathcal{Z},h_{0}\left(\cdot\right), \mf{w})$ via Cox elastic net regression, which was originally implemented in the \texttt{glmnet} R package \cite{simon2011regularization}. In our \model~implementation, we use the Python wrapper of the \texttt{glmnet} (\url{https://pypi.org/project/glmnet/}).

Upon training \model, we obtain $\widehat{\boldsymbol{\Phi}}^{(m)}$, where $m=1, \ldots, M$ indexes modalities, the hyperparameters $\widehat{\boldsymbol{\upalpha}}$ and $\widehat{\mathcal{B}}$, and the point estimates of the Cox regression coefficients $\hat{\mf{w}}$ along with the two-component mixture models trained for each PheCode-guided topic (\suppsec~\ref{sec:prior}).

\subsection{Inferring personalized survival probabilities}\label{prediction}
For a new patient $j'$ with EHR documents denoted as $x^{(m)}_{1: N^{(m)},j'}, m=1,\ldots,M$. We first compute the topic assignments by the following steps: 
\begin{enumerate}
    \item For every token $i = 1,\ldots, N^{(m)}_{j'}$ over every modality $m=1,\ldots,M$: 
        \begin{align*}
            \gamma^{(m)}_{j'ik} &\propto \left(\widehat{\alpha}_{k} \pi_{j'k} +\left[n_{j'\bullet k}^{(\bullet )}\right]_{(j',-i)}\right) \widehat{\boldsymbol{\upphi}}^{(m)}_{x^{(m)}_{j'ik}}
        \end{align*}
        where $\pi_{j'k}$ is inferred using the trained 2-component mixture model on the training data, $\widehat{\alpha}_k$ is the estimated global hyperparameter, and $\widehat{\boldsymbol{\upphi}}_{1:K}^{(m)}$ is the estimated topic distributions from the training data.
    \item Update sufficient statistics:
        \begin{align*}
            n^{(\bullet)}_{j' \bullet k} &= \sum^M_{m=1}\sum^{N^{(m)}_{j'}}_{i=1}\gamma^{(m)}_{j'ik}
        \end{align*}
    \item Evaluate the log marginal likelihood:
        \begin{align*}
            \mathcal{L}_{j'} = \sum^M_{m=1}\sum^{N^{(m)}_{j'}}_{i=1}\sum^{K}_{k=1}\log\uptheta_{j'k} \phi^{(m)}_{k x^{(m)}_{j' i}}
        \end{align*}
    where $\theta_{j'k} = n^{(\bullet)}_{j' \bullet k}/\sum^K_{k'=1}n^{(\bullet)}_{j' \bullet k'}$.
    \item Repeat 1 and 2 until 3 converges.
\end{enumerate}

Finally, we compute the mean of the variational estimates of the topic assignments:
\begin{align*}
    \bar{\gamma}_{j'k} &= \sum^M_{m=1}\frac{1}{N^{(m)}_{j'}}\sum^{N^{(m)}_{j'}}_{i=1} \gamma^{(m)}_{j'ik}    
\end{align*}


Using $\bar{\boldsymbol{\upgamma}}_{j'} ={\left[ \bar{\gamma}_{j'k} \right]}_{k=1}^K$ and survival coefficients $\widehat{\mf{w}}$, we can calculate the estimated hazard ratio for the new patient: 
    $$
        \widehat{\mathrm{HR}}_{j'} = \exp\left(\widehat{\mf{w}}^{\top} \bar{\boldsymbol{\upgamma}}_{j'}\right)
    $$ 
Moreover, we can compute the survival function for patient $j$ up to time $t$:
\begin{align*}
    P(T_{j'} > t) = \exp[-H_{0}(t)\widehat{\mathrm{HR}}_{j'}]
\end{align*}
where $H_{0}(t) = \int_0^t h_0(u) \mathrm{d} u$ denoted as the baseline cumulative hazard function. With this survival function, we can generate personalized survival curve for every patient and estimate their survival probability at specific time points \cite{cox1972regression}.

\section{Data processing and experiments}
\subsection{Simulation design 1}\label{sec:sim1}
We designed a simulation study to evaluate \model~in terms of the accuracy of (1) identifying topics associated with patient survival and (2) predicting patient survival times.
We set the vocabulary to be 1000 words ($V=1000$) and simulated 500 distinct topics ($K=500$). We sampled 8000 patients and each patient consists of 100 tokens using the data generative process of \model. 
For each patient  $j \in \{1,\ldots,8000\}$, the topic proportions $\boldsymbol{\uptheta}_j$ was sampled from a Dirichlet distribution with the hyperparameter $\boldsymbol{\alpha}$ sampled from a Gamma distribution with a shape parameter of 10 and a scale parameter of 1. 
The topic distributions $\boldsymbol{\upphi}_k$ was sampled from a $V$-dimensional Dirichlet distribution with the hyperparameter $\mathcal{B}$ sampled from a Gamma distribution with shape and scale parameters of 2 and 500, respectively. 
The topic assignment $z_{ji}\in\{1,\ldots,K\}$ was sampled from a Categorical distribution at the rate set to $\boldsymbol{\uptheta}_j$, and word assignments $x_{ji}\in\{1,\ldots,V\}$ were sampled from another Categorical distribution at the rate $\boldsymbol{\upphi}_{z_{ji}}$. 

To  evaluate whether our model can identify mortality-related topics, we set the survival coefficients $\mf{w}$ to be a sparse vector. Specifically, we set 50 out of the 450 coefficients to 6 and the rest to 0. We then computed the topic proportion $\bar{\mf{z}}_j$ for each patient $j$. 

Survival time $T_j$ were simulated via the Cox model:
    $$
        T_j = H_0^{-1}\left(-\log(U) \exp(-\mf{w}^{\top}\bar{\mf{z}}_j)\right)
    $$
where $U$ is a uniformly distributed random variable on the interval $[0, 1]$, and the transformation is done through the inverse of the baseline hazard function $H_0^{-1}$ \cite{bender2005generating}.  
We chose $H_0^{-1}$  based on the distribution of the survival times. For simplicity, we adopt the Exponential distribution, a common choice in survival analysis. In this scenario, the cumulative baseline hazard function is expressed as $H_0(t) = \lambda t$, with $\lambda$ being a hyperparameter set to 1 for simplicity, leading to the inverse function $H_0^{-1}(t) = \lambda^{-1} t$.

\subsection{Simulation design 2}\label{sec:sim2}
\hi{To create a simulated dataset that closely replicates real-world data, we focused on the CHD dataset, utilizing diagnosis codes documented prior to the first ICU discharge of CHD patients for predicting mortality time. Our simulation was tailored to mirror the CHD dataset's specific attributes, including a total of 8211 patients (P = 8211) and maintaining consistency with the actual count of phenotype topics found in the CHD dataset (K = 490).}

\hi{For the simulation of topic distributions $\boldsymbol{\upphi}_k$, we drew from a $V$-dimensional Dirichlet distribution, with each hyperparameter $\boldsymbol{\upbeta}_k$ determined by the relationship between ICD-9 codes and PheCodes. Specifically, for a given ICD-9 code $v$ and PheCode $k$, we set $\beta_{k v}=1$ if there exists a mapping between $v$ and $k$; otherwise, $\beta_{k v}=0$. To satisfy the Dirichlet distribution requirement that $\boldsymbol{\upbeta}_k > 0$, we transformed the mapping to $\beta_{k v}=\beta_{k v} \times 3+0.6$. This adjustment ensures that the simulated topic distributions closely approximate those learned from the CHD dataset by reflecting the distribution of words within a topic and highlighting dataset-specific signals and differences.}

\hi{For each patient, the topic proportions $\boldsymbol{\uptheta}_j$ were directly sampled based on the observed patient's PheCode frequency from the CHD dataset. The number of records $N_j$ for each patient was also matched to the CHD dataset to preserve information density and sparsity. We then simulated the ICD codes for the patient as follows: for each code $i$, the topic assignment $z_{j i} \in\{1, \ldots, K\}$ was sampled from a Categorical distribution parameterized by $\boldsymbol{\uptheta}_j$. The ICD code for $x_{j i} \in\{1, \ldots, V\}$ was then sampled from the Categorical distribution parameterized by $\boldsymbol{\upphi}_{z_{ji}}$.}
\hi{We then randomly designated 10\% of the Phenocode topics to have survival coefficients $w_k$ set to 6, with the remainder set to 0. Survival times $T_j$ for each patient was sampled the same way as in Simulation Design 1}


\subsection{Preprocessing of the Quebec CHD data}\label{application to CHD Dataset}

We leveraged the inpatient and outpatient ICD codes from a patient’s first documented heart failure episode to predict their subsequent time to death, measured in days. The cohort for this study was the Congenital Heart Disease (CHD) claim database. This dataset combines the Physician Services and Claims spanning from 1983 to 2010, the Hospital Discharge Summaries from 1987 to 2010, and the Vital Status records from 1983 to 2010. The dataset contains 8,211 CHD subjects, who experienced at least one heart failure and had a recorded death date. The data were constructed by collating all the ICD codes recorded during the hospitalization of the first heart failure episode. In total, there are 75,187 records and 1,767 unique ICD9 codes. We mapped the ICD9 codes to 498 unique PheCodes in one-decimal code format for the guided topic inference (\suppsec~\ref{sec:prior}). We selected PheCodes that appeared in over 25\% of the patient population within the dataset. The survival time of each CHD patient is the time difference between the death date and the discharge date of the first heart failure hospitalization.


\subsection{Preprocessing of MIMIC-III data}\label{application to the MIMIC-III Dataset}

To demonstrate the generalization of \model~and the ability for inferring multi-modal EHR topics, we made use of the Medical Information Mart for Intensive Care III (MIMIC-III) dataset \cite{johnson2016mimic}. MIMIC-III is a comprehensive dataset originating from the Beth Israel Deaconess Medical Center in Boston, MA, encompassing 53,423 distinct hospital admissions across 38,597 adult patients and 7,870 neonates from 2001 to 2012. The dataset was downloaded from the PhysioNet database (\url{mimic.physionet.org}) under its user agreement. We carried out the same preprocessing as described in \cite{ahuja2022mixehr}. 
We then selected patients who had multiple inpatient records and a documented time of death. We utilized all available EHR information up to the discharge time of the first inpatient stay to predict the time lapse the patient survived since the ICU discharge. To refine our dataset for more accurate predictions, we specifically filtered out patients whose discharge date from their first inpatient admission coincided with their date of death. The final dataset consisted of 1,458 patients, of which 1,168 were used for training the model and 290 for testing. Among these patients, there are 55,529 unique features among five EHR types including clinical notes (47,383), ICD-9 codes (3293), lab tests (588), prescriptions (3,444), and DRG codes (821).


\subsection{Evaluation}
To evaluate the \model's ability of predicting patient survival time, we utilized dynamic area under the ROC (AUC) curve \cite{uno2007evaluating, hung2010estimation, lambert2016summary}, a modification of the traditional ROC curve particularly suited for survival data analysis. Dynamic AUC extends the concept of AUC to survival data by defining time-dependent sensitivity (true positive rate) and specificity (true negative rate). In this context, cumulative cases include individuals who experienced an event by or before a specific time $\{ j \mid T_j \leq t, \; j =1,\ldots,P \}$, while cumulative controls are those for whom the event occurs after this time $\{ j \mid T_j > t, \; j =1,\ldots,P \}$. The corresponding cumulative/dynamic AUC evaluates the model's ability to distinguish between subjects who experienced an event by a given time $\left( T_j \leq t \right)$ and those who experienced it later $\left( T_j > t \right)$.

Given an estimated risk ratio $\widehat{\mathrm{HR}}_{j}$ for the $j$-th individual, the cumulative/dynamic AUC at time $t$ is defined as:
$$\widehat{\mathrm{AUC}}(t)=\frac{\sum_{i=1}^P \sum_{j=1}^P I\left(T_j>t\right) I\left(T_i \leq t\right) I\left(\widehat{\mathrm{HR}}_{j} \leq \widehat{\mathrm{HR}}_{i}\right)}{\left(\sum_{j=1}^P I\left(T_i>t\right)\right)\left(\sum_{j=1}^P I\left(T_i \leq t\right) \right)}$$
Building on this, we define a sequence of time points and calculate the cumulative/dynamic AUC at each point in this series, thereby constructing the Dynamic AUC curve.

\section{Results}\label{results}

\begin{figure}[t!]
\centering
\includegraphics[width=1.0\textwidth]{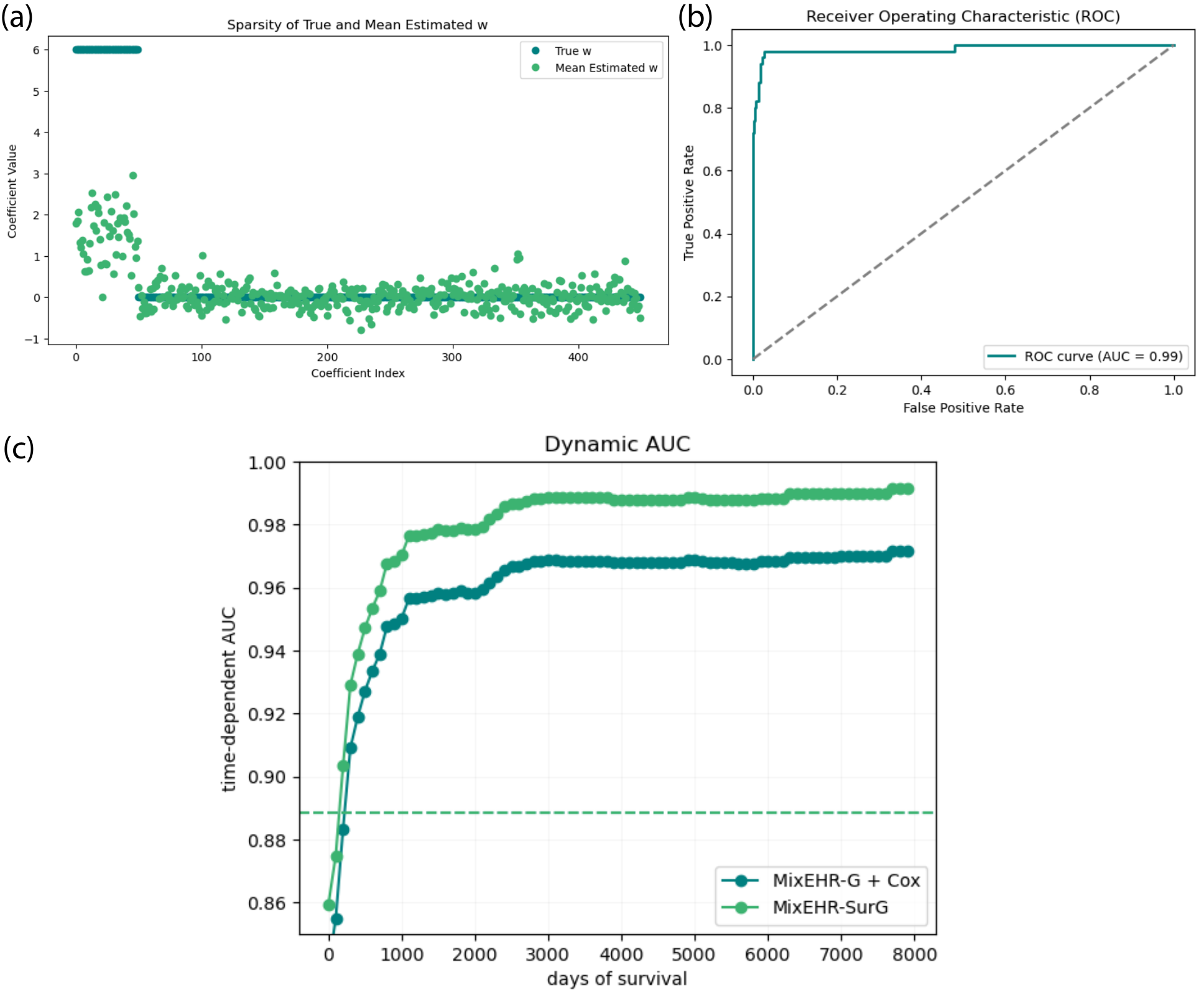}
\caption{Simulation Results for \model. (a) Scatter plot comparing the estimated coefficients $\mf{w}$ (in green) with their true values (in blue). (b) ROC curve for predicting zero coefficients. (c) Dynamic AUC curves to evaluate survival time prediction. }\label{fig:Simulation_rlt}
\end{figure}

\subsection{Simulation}

\model~demonstrates high sensitivity and specificity in detecting the 50 true survival-associated phenotypes out of the 450 phenotypes (\fig~\ref{fig:Simulation_rlt}a,b; \suppsec~\ref{appendix-eval}). Notably, the true effect size is 6, and the model estimates is between 2 and 3, which is due to the L1/2-regularization (i.e., elastic net) on $\mf{w}$ via the regularized Cox regression (\suppsec~\ref{sgd}). 
We then evaluated \model~in terms of predicting survival times in comparison to pipeline approach that ran MixEHR-G followed by Cox regression. This comparison was made using dynamic AUC curves (\fig~\ref{fig:Simulation_rlt}c), which provide a nuanced measure of sensitivity and specificity over time for survival data. \hi{\model~slightly improved over MixEHR-G with mean AUC of 0.89 versus mean AUC of 0.88, respectively. To assess whether the improvement is statistically significant, we repeated the simulations 10 times and computed the Wilcoxon signed-rank test, which yielded a p-value of 0.0488 (\supfig~\ref{fig:simbox}).}
\hi{We conducted another simulation closely based on the real-world MIMIC-III and CHD datasets (\sect~\ref{sec:sim2}). As expected, we observed lower AUCs but the relative performance between \model~and Coxnet-MixEHR-G are similar (\supfig~\ref{fig:sim2}).}

\begin{figure}[t!]
\centering
\includegraphics[width=\textwidth]{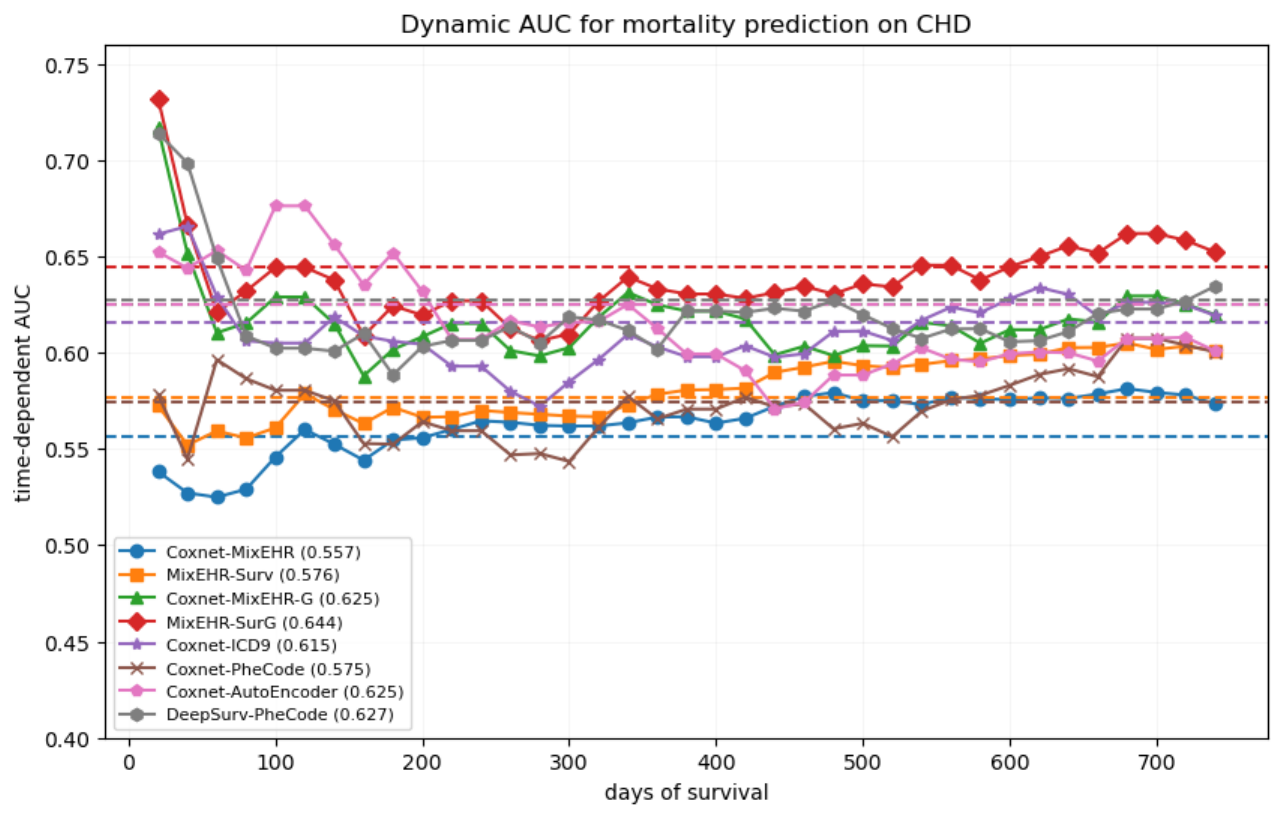}
\caption{Dynamic AUC curves for predicting time to death in CHD patients. We built a series of time points starting from 20 and incrementing by 20 up to 755. For each of these time points, we computed the cumulative AUC, which then formed the Dynamic AUC curve. \hi{The mean AUC over time for each method was indicated as dash lines and in the bracket after each method in the legend. The compared methods are:
    \textbf{Coxnet-MixEHR}: A pipeline approach by training MixEHR first and then training a Cox elastic net (Coxnet) using the topic mixture from MixEHR as the input features; 
    \textbf{MixEHR-Surv}: MixEHR with the Cox supervision but without the phecode guided prior for the topic inference; 
    \textbf{Coxnet-MixEHR-G}: A pipeline approach by training MixEHR-G first and then training a Cox elastic net (Coxnet) using the topic mixture from MixEHR-G as the input features;
    \textbf{MixEHR-SurG}: the proposed method in this paper; 
    \textbf{Coxnet-ICD9}: Cox elastic net (Coxnet) using ICD9 code as input features; 
    \textbf{Coxnet-PheCode}: Coxnet using PheCode as input features; 
    \textbf{Coxnet-AutoEncoder}: Coxnet using the output of an autoencoder as input features; 
    \textbf{DeepSurv-PheCode}: Deep survival model using PheCode as input features.
} 
}\label{fig:CHD_rlt1}
\end{figure}

\begin{figure}[t!]
\centering
\includegraphics[width=0.9\textwidth]{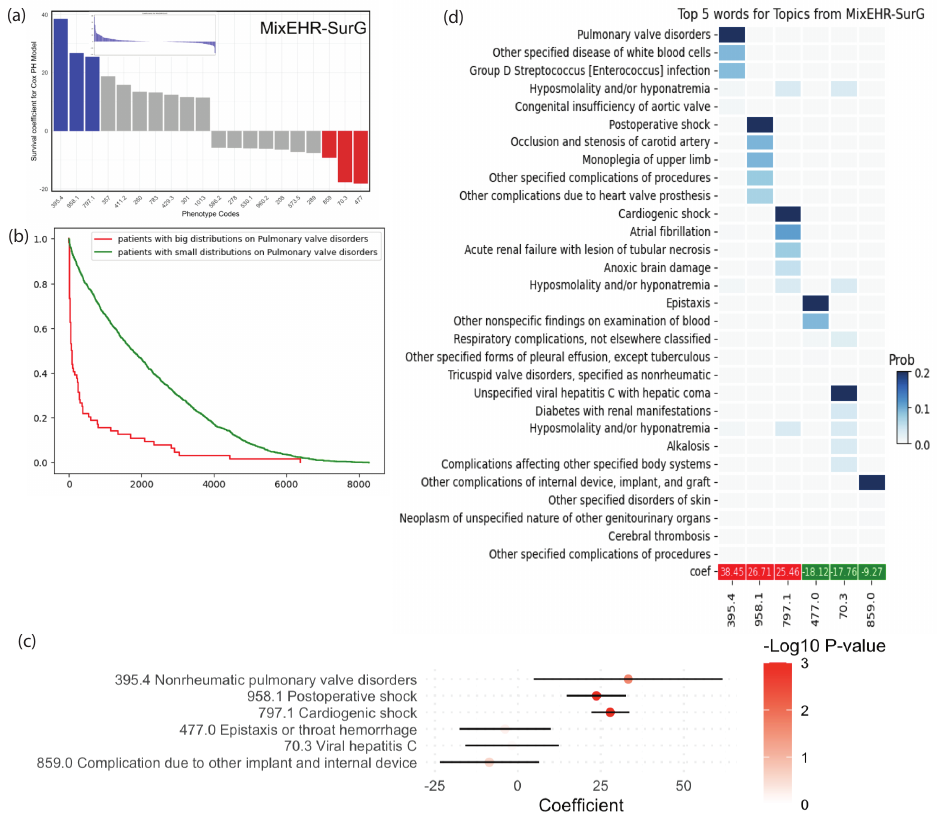}
\caption{Mortality-related phenotypes for CHD patients who experienced first heart failure hospitalization.
(a) Bar plot of the survival regression coefficients $\mf{w}$. The effect size of the 10 most positive and the 10 most negative phenotypes are displyaed as barplot. The positive value refers to phenotypes that are associated with high risk of mortality and the negative value refers to phenotypes associated with low mortality risk. The inset at the up-left corner contains the bar plot for all the estimated $w_k, k =1,\ldots,K$ ranked from the largest value to the smallest value. The top 3 and bottom phenotypes were colored in blue and red, respectively.
(b) The survival curves of patient with high and low risk of nonrheumatic pulmonary valve disorder (NPVD) (395.4). Patients were divided into two groups based on their topic proportions. The red curve represents patients with a higher topic proportion (top 30\%) in NPVD as shown by a significantly steeper decline and lower survival probability over time. The green curve, representing patients with lower topic proportions of NPVD phenotype, shows a more gradual decline, reflecting a comparatively lower risk of mortality. 
(c) Effect size of the mortality-related phenotypes. We ran simple Cox regression per phenotype topic to obtain their marginal effect size and 95\% confidence interval of the  top 3 high risk and bottom 3 low risk mortality-associated phenotypes as identified by \model~in panel (a). Points indicate the coefficient values, Error bars show the 95\% confidence intervals, and colors represent the significance levels of these coefficients. 
(d) Heatmap featuring the top ICD-9 codes from the three most positively predictive and three most negatively predictive phenotypes as determined by from \model. The intensity of the colors indicates the topic probability in under each topic. The magnitude of the Cox coefficients are displayed in the last row.
}\label{fig:CHD_rlt2}
\end{figure}

\subsection{Application to the CHD dataset}\label{result for CHD}

We evaluated the survival models on CHD patient survival time predictions after their initial HF hospitalization. \model~conferred the highest mean AUC (0.645) compared to MixEHR-G with the Coxnet pipeline (0.623), MixEHR-Surv (0.576), and MixEHR with the Coxnet pipeline (0.556) and DeepSurv (0.64) (\fig~\ref{fig:CHD_rlt1}). \hihi{To further ascertain the benefit of joint topic inference and survival regression, we sampled from the test patients with replacement 10,000 times and calculated the difference of the mean AUCs for each bootstrap between \model~and MixEHR-G+Coxnet: $\Delta$AUC = AUC(MixEHR-SurG) - AUC(MixEHR-G+Coxnet). We used the 10,000 $\Delta$AUCs to construct an empirical distribution of the performance difference between the two methods. The 75\% confidence intervals (CIs) of the empirical distribution is [0.00364  0.0370], corresponding to the 12.5\% quantile and 87.5\% quantile of the empirical $\Delta$AUC distribution, respectively. Furthermore, 9,260 out of the 10,000 bootstrap $\Delta$AUC are positive, which is statistically significant at the p-value $<$ 2.5e-324 based on one-sided Binomial test with the null hypothesis being at the equal chance of producing positive and negative $\Delta$AUC over the 10,000 bootstrap samples.}

Based on the Cox regression learned simultaneously by \model, we identified the most predictive phenotypes of the post-HF survival time (\fig~\ref{fig:CHD_rlt2}a). Among these phenotypes, nonrheumatic pulmonary valve disorder (NPVD) (395.4) is the most prominent phenotype. Indeed, the CHD subjects who exhibit high topic proportion for NPVD tend to have much shorter survival time compared to the rest of the CHD subjects (\fig~\ref{fig:CHD_rlt2}b; \textbf{\ref{appendix-surv}}). We then obtained the $p$-values and confidence intervals of the six phenotypes selected for their large absolute value of $w_k$, through a Cox proportional hazards model \cite{cox1972regression}. Based on the results (\fig~\ref{fig:CHD_rlt2}c), it is evident that ``Nonrheumatic pulmonary valve disorders", ``Postoperative shock", and ``Cardiogenic shock" have emerged as significant factors contributing to the occurrence of mortality. These phenotypes are characterized by substantial positive coefficients and statistically significant $p$-values, underscoring their strong association with increased risk. Interestingly, ``Complication due to other implant and internal device" (859.0) is associated with longer survival time, which perhaps imply the deficiency of healthcare among those high risk patient group.
We then examined the underlying top ICD9 codes under the predictive phenotype topics (\fig~\ref{fig:CHD_rlt2}d). In particular, topic for NPVD includes several cardiac-related ICD codes with pulmonary valve disorders being the most prominent one as expected. Phenotype topics ``Postoperative shock" (958.1) and``Cardiogenic shock" (797.1) were also associated with the relevant ICD-9 codes, implying high topic coherence. The 3 negative topics are not heart-specific but nonetheless semantically coherent. \hi{We further validated the topic coherence based on the mutual information (MI) between the top ICD codes for the top 6 survival phenotype topics (\supfig~\ref{fig:chd_top_mi}). Indeed, we observe a clear $5\times5$ block pattern corresponding to the top 5 ICD codes for the corresponding phenotype topic along the diagonal of the MI matrix. Furthermore, the top ICD codes that are not part of the PheCode definition exhibit high MI with the PheCode-defining ICD code, implying that they are not only related to the phenotype but also co-occur with the PheCode-defining ICD code in the actual patient records. This also suggests that \model~does not completely rely on the PheCode guide not also driven by the CHD data in characterizing the phenotype topic distributions.}

\begin{figure}[t!]
\centering
\includegraphics[width=0.75\textwidth]{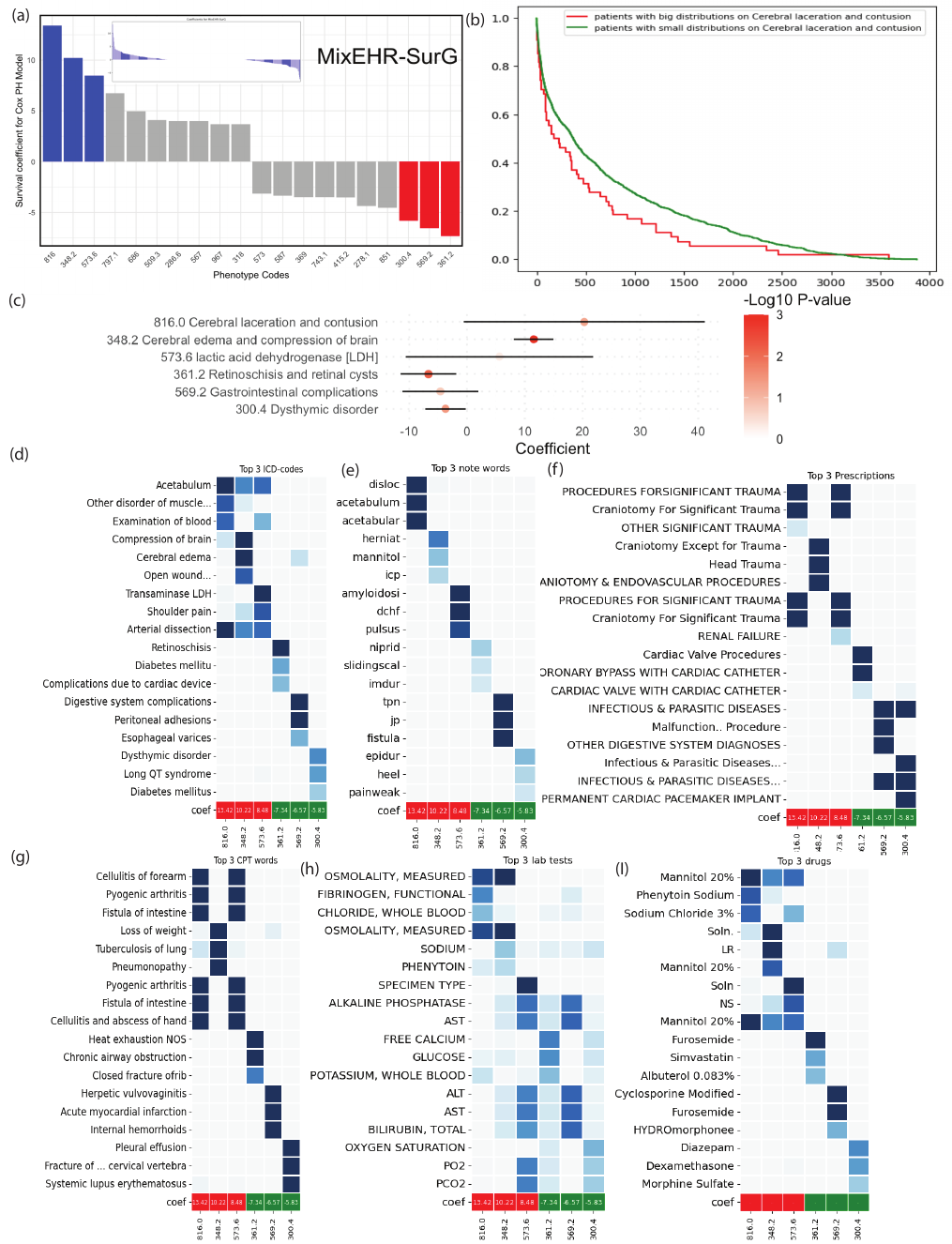}
\caption{\scriptsize 
Mortality-related multi-modal phenotypes for the ICU patients from MIMIC-III dataset. 
(a) Bar plot of survival regression coefficients $\mathbf{w}$. Inset in the upper-left corner that displays all estimated coefficients $w_k$ for $k =1,\ldots,K$, organized from the largest to the smallest. We highlighted the top 10 coefficients with the largest effect sizes of positive and negative values. The positive coefficients are linked to phenotypes that elevate the risk of mortality, while the  negative coefficients are associated with phenotypes that are predictive of lower mortality risk. The blue and red color highlight the top 3 and bottom 3 phenotypes, respectively, that we analyzed in-depth below.
(b) Survival curves delineating two patient groups based on their distribution levels within the ``Cerebral laceration and contusion" topic, which is the most significant predictor of high mortality risk. The red curve illustrates patients with a higher distribution in this topic, exhibiting a more pronounced decline in survival; the green curve, indicative of patients with a lower distribution, depicts a more gradual decrease in survival, pointing to a lower mortality risk.  
(c) Effect size of the mortality-related phenotypes. We ran simple Cox regression per phenotype topic to obtain their marginal effect size and 95\% confidence interval of the  top 3 high risk and bottom 3 low risk mortality-associated phenotypes as identified by \model~in panel (a). Points indicate the coefficient values, Error bars show the 95\% confidence intervals, and colors represent the significance levels of these coefficients.
(d)-(I) Heatmap showing the top ICD-9 codes, clinical note terms, CPT descriptors, medications, and lab tests under the moratality-related phenotypes. The color gradation indicate the prevalence of each feature within each phenotype topic. The last row indicates the Cox regression coefficients.}\label{fig:Mimic_rlt2}
\end{figure}

\subsection{Application to MIMIC-III dataset}\label{result for mimic}

We then benchmarked each method on the mortality prediction using the MIMIC-III dataset (\supfig~\ref{fig:Mimic_rlt1}). Among the four model variants, \model~achieved the highest mean AUC (0.54), closely followed by the Coxnet-MixEHR-G pipeline (0.53). MixEHR-Surv and the Coxnet-MixEHR pipeline conferred mean AUCs of 0.48 and 0.39, respectively. 
\hihi{Moreover, MixEHR-SurG significantly outperformed the runner up  baseline MixEHR-G+Coxnet with the 75\% CI estimated from 10,000 bootstrap equal to [0.000913, 0.0360] (\supfig~\ref{fig:bootstrap_boxplot}b), and 8,967 out of the 10,000 bootstrap $\Delta$AUC = AUC(MixEHR-SurG) - AUC(MixEHR-G+Coxnet) being positive (p-value $<$2.5e-324 based on one-sided Binomial test rejecting the null hypothesis at the equal chance of getting positive and negative $\Delta$AUC)}.

Nonetheless, the absolute AUC level is lower than the CHD data, which may be due to the smaller sample size and more diverse causes of death. In addition, the relationship between patient data and survival outcomes in MIMIC-III is influenced by the dataset's heterogeneity and the emergency nature of many admissions, where acute conditions can overshadow chronic illness history in predicting mortality. The CHD dataset, by contrast, lends itself to more accurate predictions due to the focused nature of the cohort. Patients with CHD often have extensive medical histories and a narrower range of complications, providing a stronger and more direct signal for predicting mortality.

We then sought to identify phenotype topics that are indicative of the short-term mortality based on the Cox regression coefficients that were jointly fit with the EHR data by our \model~(\fig~\ref{fig:Mimic_rlt2}a). The most prominent phenotype topic is ``Cerebral laceration and contusion (816.0)", which also separates patients into high and low risk groups (\fig~\ref{fig:Mimic_rlt2}b). We subsequently assessed the $p$-values and confidence intervals of six phenotypes with the largest absolute values of $w_k$, through a Cox proportional hazards model \cite{cox1972regression} (\fig~\ref{fig:Mimic_rlt2}c). The results confirm the significance of ``Cerebral laceration and contusion", ``Cerebral edema and compression of brain" and ``Dysthymic disorder". Specifically, ``Cerebral laceration and contusion" and ``Cerebral edema and compression of brain" show a positive correlation with an increased risk of mortality, while ``Dysthymic disorder" indicates a negative correlation, suggesting a potential protective effect against mortality. Indeed, traumatic brain injuries often lead to servere morbidity and ultimately death. 
Conversely, \model~reveals conditions such as ``Retinoschisis and retinal cysts" and ``Dysthymic disorder" with large negative survival coefficients, suggesting a low immediate threat to life. Retinoschisis and retinal cysts typically do not directly impact survival unless complicated by additional factors, and ``Dysthymic disorder" while affecting quality of life, generally does not shorten life expectancy in the absence of other comorbid conditions.

\hi{We further performed Kaplan-Meier (KM) survival analysis and computed the p-values using one-sided log-rank tests for the top ICD codes under the top 6 survival phenotype topics (\supfig~\ref{fig:mimic-icd-heatmap}a). We observe that the top codes associated with the first three phenotypes, which have higher survival coefficients, display significant marginal effects of increased hazard risks. Conversely, the top codes linked to the last three phenotypes exhibit significant marginal effects of reduced hazard risks. These findings confirm that our model can effectively pinpoint terms with substantial impacts on survival. Some ICD-9 codes, such as '8080 Closed fracture of acetabulum' and '72889 Disorder of muscle, ligament, and fascia', are not significant by themselves but contribute in aggregate to the survival phenotype suhc as `816.0'.
Under the topics of traumatic brain injuries, namely ``816.0: Cerebral laceration and contusion" and ``348.2: Cerebral edema and compression of brain", the top ICD codes are semantically coherent (\fig~\ref{fig:Mimic_rlt2}d). Quantitatively, we computed the mutual information (MI) between the top ICD codes (\supfig~\ref{fig:mimic-icd-heatmap}b). As expected, the top codes under the same phenotype topic exhibit high MI, implying a high topic coherence. This may not be surprising as some of the ICD codes were used to define the PheCode, which were then used to build the topic prior.}

To gain further insights to these topics, we examined the top EHR codes from non-ICD modality topics (\fig~\ref{fig:Mimic_rlt2}e-I).
In clinical notes (\fig~\ref{fig:Mimic_rlt2}e), we identified top words related to mannitol, a diuretic used to reduce intracranial pressure \cite{wakai2005mannitol}. \hi{Mannitol is the treatment of cerebral edema (accumulation of excessive fluid in the brain). The fact that it is the top drug code for the top risk mortality phenotype 816.0 suggests the severity of the condition.} Indeed, we also found ``Osmolality, Measured" to be the top term under the laboratory modality and ``Mannitol 20\%" as the top term under the same topic from prescription modality (\fig~\ref{fig:Mimic_rlt2}I).
In addition, the DRG (Diagnosis-Related Group) topic modality exhibit strong connection with the ICD-modality topic despite the fact that DRG codes are not part of the PheCode definition. \hi{Most of these top codes also exhibit consistently significant marginal effect size based on the KM test and coherence in terms of mutual information (\supfig~\ref{fig:mimic-drug-heatmap}, \ref{fig:mimic-DRG-heatmap}, \ref{fig:mimic-lab-heatmap}, \ref{fig:mimic-cpt-heatmap}).}
Together, these results showcases the \model's ability to harness non-ICD modality to enrich the phenotyping, which is consistent to what we observed in MixEHR-G \cite{ahuja2022mixehr}. 

Our results suggested that brain injuries are the strong mortality-indicators and the topic coherence across modalities provide the fine-grained markers for screening high-risk patients in the future.

\section{Discussion}\label{discussion}

Effective utilization of EHR data holds the promise to automate phenotyping \cite{alzoubi2019review} and identifying prognosis markers \cite{yuan2021performance}. \model~extends EHR topic modeling to  survival topic model with the identifiable topics by utilizing the patient survival time and PheCode definitions, respectively. We demonstrated the utility of \model~via both simulation and real-world EHR data including the Quebec CHD and MIMIC-III datasets \cite{johnson2016mimic}. The results from these rigorous experiments highlights our contribution in \model~as an effective approach to identify clinically meaningful phenotypes that implicate mortality. 

Despite this advance, there are several limitations in our method. First, EHR data often contain hierarchical structures of phenotypes that are yet to be unraveled. For instance, leveraging advanced hierarchical topic modeling \cite{baytas2016phenotree, pivovarov2015learning} could shed light on sub-phenotypes and their interactions within broader disease categories. While we have harnessed cross-sectional data effectively, the longitudinal nature of EHRs, characterized by patient trajectories and time-stamped health events, presents an opportunity to explore temporal patterns and trends \cite{defossez2014temporal} in future studies.

Although \model~showcases predictive prowess, it does not have the same level of expressiveness as deep neural networks. The integration of deep learning with topic model \cite{bhat2020deep, cao2015novel} could potentially enhance predictive performance by capturing non-linear relationships and complex interaction effects within EHR data \cite{zhao2021topic,wang2022graph,zou2022modeling,song2022automatic}. Furthermore, the challenge in distinguishing between high-mortality risk phenotypes and confounding factors remains and calls for causality-driven models \cite{veitch2020adapting, kim2013mining, rosa2011inferring, corbin2018formalising}. Future study will be dedicated to not only predict outcomes but also discern the underlying causal mechanisms, offering a more granular understanding of patient risk profiles. Causal inference that discern direct from indirect influences of phenotypes on survival outcomes will bring a step closer to effective clinical interventions.

In summary, \model~is a novel topic model that leverages EHR data for both interpretive and predictive modeling of patient survival outcomes. By successfully mapping EHR data to relevant phenotypes and delineating those with high mortality risks, \model~serves as a prototype for future systems that could offer nuanced insights into patient care. The current study lays the groundwork for subsequent research that could incorporate hierarchical data structures and temporal dynamics within EHRs \cite{baytas2016phenotree, pivovarov2015learning, defossez2014temporal}, potentially utilizing advanced machine learning techniques such as deep learning and causal inference models \cite{bhat2020deep, cao2015novel, veitch2020adapting, kim2013mining, rosa2011inferring, corbin2018formalising}. Such developments could further refine the precision of survival predictions and enhance the interpretability of complex healthcare data, ultimately leading to more informed and personalized medical decision-making. As the field advances, we anticipate that the integration of these sophisticated methodologies will yield models that not only predict but also disentangle the intricate network of disease causality within patient health trajectories \cite{do2022network, han2021disease, oh2021computational}.


\section{Ethic}
The use of the CHD data for this research was approved by the McGill University Health Centre Research Ethics Board.

\section{Acknowledgements}
Dr. Yue Li is supported by Canada Research Chair (Tier 2) in Machine Learning for Genomics and Healthcare, Natural Sciences and Engineering Research Council (NSERC) Discovery Grant (RGPIN-2016-05174).
Dr. Archer Yi Yang is supported by Natural Sciences and Engineering Research Council (NSERC) Discovery Grant (RGPIN-2019-0621). Both Dr. Yang and Dr. Li are by the FRQNT Team Research Project Grant (FRQ-NT 327788).
Dr. Ariane Marelli is supported by the Heart and Stroke Foundation grant and award (John Day, MD, Excellence Award for Heart Failure Trajectory Along the Care Continuum for Congenital Heart Disease Patients Across the Lifespan).

\section{Author contributions}
Y.L., Y.X.L., and A.Y. conceived the study. Y.X.L. Y.L., A.Y., developed the methodology. Y.X.L. implemented the software and conducted the experiments under supervision of Y.L. and A.Y. 
Y.X.L, Y.L., and A.Y. analyzed the results. Y.X.L., Y.L. and A.Y. wrote the initial draft of the manuscript. A.M. provided the CHD data. All of the authors reviewed and wrote the final version of the manuscript.





\clearpage


 \bibliographystyle{elsarticle-num}
 \bibliography{main}

\begin{thebibliography}{10}
\expandafter\ifx\csname url\endcsname\relax
  \def\url#1{\texttt{#1}}\fi
\expandafter\ifx\csname urlprefix\endcsname\relax\def\urlprefix{URL }\fi
\expandafter\ifx\csname href\endcsname\relax
  \def\href#1#2{#2} \def\path#1{#1}\fi

\bibitem{jiang2023pre}
J.~Jiang, K.~Qi, G.~Bai, K.~Schulman, Pre-pandemic assessment: a decade of progress in electronic health record adoption among us hospitals, Health Affairs Scholar 1~(5) (2023) qxad056.

\bibitem{smoller2018use}
J.~W. Smoller, The use of electronic health records for psychiatric phenotyping and genomics, American Journal of Medical Genetics Part B: Neuropsychiatric Genetics 177~(7) (2018) 601--612.

\bibitem{alzoubi2019review}
H.~Alzoubi, R.~Alzubi, N.~Ramzan, D.~West, T.~Al-Hadhrami, M.~Alazab, A review of automatic phenotyping approaches using electronic health records, Electronics 8~(11) (2019) 1235.

\bibitem{shivade2014review}
C.~Shivade, P.~Raghavan, E.~Fosler-Lussier, P.~J. Embi, N.~Elhadad, S.~B. Johnson, A.~M. Lai, A review of approaches to identifying patient phenotype cohorts using electronic health records, Journal of the American Medical Informatics Association 21~(2) (2014) 221--230.

\bibitem{jensen2012mining}
P.~B. Jensen, L.~J. Jensen, S.~Brunak, Mining electronic health records: towards better research applications and clinical care, Nature Reviews Genetics 13~(6) (2012) 395--405.

\bibitem{jensen2017analysis}
K.~Jensen, C.~Soguero-Ruiz, K.~Oyvind~Mikalsen, R.-O. Lindsetmo, I.~Kouskoumvekaki, M.~Girolami, S.~Olav~Skrovseth, K.~M. Augestad, Analysis of free text in electronic health records for identification of cancer patient trajectories, Scientific reports 7~(1) (2017) 46226.

\bibitem{javaid2022significance}
M.~Javaid, A.~Haleem, R.~P. Singh, R.~Suman, S.~Rab, Significance of machine learning in healthcare: Features, pillars and applications, International Journal of Intelligent Networks 3 (2022) 58--73.

\bibitem{miotto2016deep}
R.~Miotto, L.~Li, B.~A. Kidd, J.~T. Dudley, Deep patient: an unsupervised representation to predict the future of patients from the electronic health records, Scientific reports 6~(1) (2016) 1--10.

\bibitem{ranganath2016deep}
R.~Ranganath, A.~Perotte, N.~Elhadad, D.~Blei, Deep survival analysis, in: Machine Learning for Healthcare Conference, PMLR, 2016, pp. 101--114.

\bibitem{lee2018deephit}
C.~Lee, W.~Zame, J.~Yoon, M.~Van Der~Schaar, Deephit: A deep learning approach to survival analysis with competing risks, in: Proceedings of the AAAI conference on artificial intelligence, Vol.~32, 2018.

\bibitem{shin2021machine}
S.~Shin, P.~C. Austin, H.~J. Ross, H.~Abdel-Qadir, C.~Freitas, G.~Tomlinson, D.~Chicco, M.~Mahendiran, P.~R. Lawler, F.~Billia, et~al., Machine learning vs. conventional statistical models for predicting heart failure readmission and mortality, ESC heart failure 8~(1) (2021) 106--115.

\bibitem{cox1972regression}
D.~R. Cox, Regression models and life-tables, Journal of the Royal Statistical Society: Series B (Methodological) 34~(2) (1972) 187--202.

\bibitem{li2002kernel}
H.~Li, Y.~Luan, Kernel cox regression models for linking gene expression profiles to censored survival data, in: Biocomputing 2003, World Scientific, 2002, pp. 65--76.

\bibitem{ishwaran2008random}
H.~Ishwaran, U.~B. Kogalur, E.~H. Blackstone, M.~S. Lauer, Random survival forests (2008).

\bibitem{tibshirani1997lasso}
R.~Tibshirani, The lasso method for variable selection in the cox model, Statistics in medicine 16~(4) (1997) 385--395.

\bibitem{lundberg2017unified}
S.~M. Lundberg, S.-I. Lee, A unified approach to interpreting model predictions, Advances in neural information processing systems 30 (2017).

\bibitem{chen2022explaining}
H.~Chen, S.~M. Lundberg, S.-I. Lee, Explaining a series of models by propagating shapley values, Nature communications 13~(1) (2022) 4512.

\bibitem{lundberg2020local}
S.~M. Lundberg, G.~Erion, H.~Chen, A.~DeGrave, J.~M. Prutkin, B.~Nair, R.~Katz, J.~Himmelfarb, N.~Bansal, S.-I. Lee, From local explanations to global understanding with explainable ai for trees, Nature machine intelligence 2~(1) (2020) 56--67.

\bibitem{blei2003latent}
D.~M. Blei, A.~Y. Ng, M.~I. Jordan, Latent dirichlet allocation, Journal of machine Learning research 3~(Jan) (2003) 993--1022.

\bibitem{li2020inferring}
Y.~Li, P.~Nair, X.~H. Lu, Z.~Wen, Y.~Wang, A.~A.~K. Dehaghi, Y.~Miao, W.~Liu, T.~Ordog, J.~M. Biernacka, et~al., Inferring multimodal latent topics from electronic health records, Nature communications 11~(1) (2020) 2536.

\bibitem{song2021supervised}
Z.~Song, X.~S. Toral, Y.~Xu, A.~Liu, L.~Guo, G.~Powell, A.~Verma, D.~Buckeridge, A.~Marelli, Y.~Li, Supervised multi-specialist topic model with applications on large-scale electronic health record data, in: Proceedings of the 12th ACM Conference on Bioinformatics, Computational Biology, and Health Informatics, 2021, pp. 1--26.

\bibitem{song2022automatic}
Z.~Song, Y.~Hu, A.~Verma, D.~L. Buckeridge, Y.~Li, Automatic phenotyping by a seed-guided topic model, in: Proceedings of the 28th ACM SIGKDD Conference on Knowledge Discovery and Data Mining, 2022, pp. 4713--4723.

\bibitem{ahuja2022mixehr}
Y.~Ahuja, Y.~Zou, A.~Verma, D.~Buckeridge, Y.~Li, Mixehr-guided: A guided multi-modal topic modeling approach for large-scale automatic phenotyping using the electronic health record, Journal of biomedical informatics 134 (2022) 104190.

\bibitem{dawson2012survival}
J.~A. Dawson, C.~Kendziorski, Survival-supervised latent dirichlet allocation models for genomic analysis of time-to-event outcomes, arXiv preprint arXiv:1202.5999 (2012).

\bibitem{teh2006collapsed}
Y.~Teh, D.~Newman, M.~Welling, A collapsed variational bayesian inference algorithm for latent dirichlet allocation, Advances in neural information processing systems 19 (2006).

\bibitem{wei2017evaluating}
W.-Q. Wei, L.~A. Bastarache, R.~J. Carroll, J.~E. Marlo, T.~J. Osterman, E.~R. Gamazon, N.~J. Cox, D.~M. Roden, J.~C. Denny, Evaluating phecodes, clinical classification software, and icd-9-cm codes for phenome-wide association studies in the electronic health record, PloS one 12~(7) (2017) e0175508.

\bibitem{simon2011regularization}
N.~Simon, J.~Friedman, T.~Hastie, R.~Tibshirani, Regularization paths for cox’s proportional hazards model via coordinate descent, Journal of statistical software 39~(5) (2011) 1.

\bibitem{bender2005generating}
R.~Bender, T.~Augustin, M.~Blettner, Generating survival times to simulate cox proportional hazards models, Statistics in medicine 24~(11) (2005) 1713--1723.

\bibitem{johnson2016mimic}
A.~E. Johnson, T.~J. Pollard, L.~Shen, L.-w.~H. Lehman, M.~Feng, M.~Ghassemi, B.~Moody, P.~Szolovits, L.~Anthony~Celi, R.~G. Mark, Mimic-iii, a freely accessible critical care database, Scientific data 3~(1) (2016) 1--9.

\bibitem{uno2007evaluating}
H.~Uno, T.~Cai, L.~Tian, L.-J. Wei, Evaluating prediction rules for t-year survivors with censored regression models, Journal of the American Statistical Association 102~(478) (2007) 527--537.

\bibitem{hung2010estimation}
H.~Hung, C.-T. Chiang, Estimation methods for time-dependent auc models with survival data, Canadian Journal of Statistics 38~(1) (2010) 8--26.

\bibitem{lambert2016summary}
J.~Lambert, S.~Chevret, Summary measure of discrimination in survival models based on cumulative/dynamic time-dependent roc curves, Statistical methods in medical research 25~(5) (2016) 2088--2102.

\bibitem{wakai2005mannitol}
A.~Wakai, I.~G. Roberts, G.~Schierhout, Mannitol for acute traumatic brain injury, Cochrane Database of Systematic Reviews~(4) (2005).

\bibitem{yuan2021performance}
Q.~Yuan, T.~Cai, C.~Hong, M.~Du, B.~E. Johnson, M.~Lanuti, T.~Cai, D.~C. Christiani, Performance of a machine learning algorithm using electronic health record data to identify and estimate survival in a longitudinal cohort of patients with lung cancer, JAMA Network Open 4~(7) (2021) e2114723--e2114723.

\bibitem{baytas2016phenotree}
I.~M. Baytas, K.~Lin, F.~Wang, A.~K. Jain, J.~Zhou, Phenotree: Interactive visual analytics for hierarchical phenotyping from large-scale electronic health records, IEEE Transactions on Multimedia 18~(11) (2016) 2257--2270.

\bibitem{pivovarov2015learning}
R.~Pivovarov, A.~J. Perotte, E.~Grave, J.~Angiolillo, C.~H. Wiggins, N.~Elhadad, Learning probabilistic phenotypes from heterogeneous ehr data, Journal of biomedical informatics 58 (2015) 156--165.

\bibitem{defossez2014temporal}
G.~Defossez, A.~Rollet, O.~Dameron, P.~Ingrand, Temporal representation of care trajectories of cancer patients using data from a regional information system: an application in breast cancer, BMC medical informatics and decision making 14~(1) (2014) 1--15.

\bibitem{bhat2020deep}
M.~R. Bhat, M.~A. Kundroo, T.~A. Tarray, B.~Agarwal, Deep lda: A new way to topic model, Journal of Information and Optimization Sciences 41~(3) (2020) 823--834.

\bibitem{cao2015novel}
Z.~Cao, S.~Li, Y.~Liu, W.~Li, H.~Ji, A novel neural topic model and its supervised extension, in: Proceedings of the AAAI Conference on Artificial Intelligence, Vol.~29, 2015.

\bibitem{zhao2021topic}
H.~Zhao, D.~Phung, V.~Huynh, Y.~Jin, L.~Du, W.~Buntine, Topic modelling meets deep neural networks: A survey, arXiv preprint arXiv:2103.00498 (2021).

\bibitem{wang2022graph}
Y.~Wang, R.~Benavides, L.~Diatchenko, A.~V. Grant, Y.~Li, A graph-embedded topic model enables characterization of diverse pain phenotypes among uk biobank individuals, Iscience 25~(6) (2022).

\bibitem{zou2022modeling}
Y.~Zou, A.~Pesaranghader, Z.~Song, A.~Verma, D.~L. Buckeridge, Y.~Li, Modeling electronic health record data using an end-to-end knowledge-graph-informed topic model, Scientific Reports 12~(1) (2022) 17868.

\bibitem{veitch2020adapting}
V.~Veitch, D.~Sridhar, D.~Blei, Adapting text embeddings for causal inference, in: Conference on Uncertainty in Artificial Intelligence, PMLR, 2020, pp. 919--928.

\bibitem{kim2013mining}
H.~D. Kim, M.~Castellanos, M.~Hsu, C.~Zhai, T.~Rietz, D.~Diermeier, Mining causal topics in text data: iterative topic modeling with time series feedback, in: Proceedings of the 22nd ACM international conference on information \& knowledge management, 2013, pp. 885--890.

\bibitem{rosa2011inferring}
G.~J. Rosa, B.~D. Valente, G.~de~los Campos, X.-L. Wu, D.~Gianola, M.~A. Silva, Inferring causal phenotype networks using structural equation models, Genetics Selection Evolution 43~(1) (2011) 1--13.

\bibitem{corbin2018formalising}
L.~J. Corbin, V.~Y. Tan, D.~A. Hughes, K.~H. Wade, D.~S. Paul, K.~E. Tansey, F.~Butcher, F.~Dudbridge, J.~M. Howson, M.~W. Jallow, et~al., Formalising recall by genotype as an efficient approach to detailed phenotyping and causal inference, Nature Communications 9~(1) (2018) 711.

\bibitem{do2022network}
I.~F. do~Valle, B.~Ferolito, H.~Gerlovin, L.~Costa, S.~Demissie, F.~Linares, J.~Cohen, D.~R. Gagnon, J.~M. Gaziano, E.~Begoli, et~al., Network-medicine framework for studying disease trajectories in us veterans, Scientific Reports 12~(1) (2022) 12018.

\bibitem{han2021disease}
X.~Han, C.~Hou, H.~Yang, W.~Chen, Z.~Ying, Y.~Hu, Y.~Sun, Y.~Qu, L.~Yang, U.~A. Valdimarsd{\'o}ttir, et~al., Disease trajectories and mortality among individuals diagnosed with depression: a community-based cohort study in uk biobank, Molecular psychiatry 26~(11) (2021) 6736--6746.

\bibitem{oh2021computational}
W.~Oh, M.~S. Steinbach, M.~R. Castro, K.~A. Peterson, V.~Kumar, P.~J. Caraballo, G.~J. Simon, A computational method for learning disease trajectories from partially observable ehr data, IEEE journal of biomedical and health informatics 25~(7) (2021) 2476--2486.

\bibitem{liao2019high}
K.~P. Liao, J.~Sun, T.~A. Cai, N.~Link, C.~Hong, J.~Huang, J.~E. Huffman, J.~Gronsbell, Y.~Zhang, Y.-L. Ho, et~al., High-throughput multimodal automated phenotyping (map) with application to phewas, Journal of the American Medical Informatics Association 26~(11) (2019) 1255--1262.

\bibitem{griffiths2004finding}
T.~L. Griffiths, M.~Steyvers, Finding scientific topics, Proceedings of the National academy of Sciences 101~(suppl\_1) (2004) 5228--5235.

\bibitem{sato2012rethinking}
I.~Sato, H.~Nakagawa, Rethinking collapsed variational bayes inference for lda, arXiv preprint arXiv:1206.6435 (2012).

\bibitem{minka2000estimating}
T.~Minka, Estimating a dirichlet distribution (2000).

\bibitem{sksurv}
S.~P{\"o}lsterl, \href{http://jmlr.org/papers/v21/20-729.html}{scikit-survival: A library for time-to-event analysis built on top of scikit-learn}, Journal of Machine Learning Research 21~(212) (2020) 1--6.
\newline\urlprefix\url{http://jmlr.org/papers/v21/20-729.html}

\bibitem{survival-package}
T.~M. Therneau, \href{https://CRAN.R-project.org/package=survival}{A Package for Survival Analysis in R}, r package version 3.5-7 (2023).
\newline\urlprefix\url{https://CRAN.R-project.org/package=survival}

\end{thebibliography}






\makeatletter
\renewcommand{\thesection}{S\@arabic\c@section}
\renewcommand{\thetable}{S\@arabic\c@table}
\renewcommand{\thefigure}{S\@arabic\c@figure}

\setcounter{section}{0}
\setcounter{figure}{0}
\setcounter{table}{0}

\clearpage

\vbox{%
    \centering
    \rule{\textwidth}{4pt} \relax
    \vskip 0.19in
    \vskip -5.5pt
    \large{\textbf{Supplementary Materials\\\model: a joint proportional hazard and guided topic model for inferring mortality-associated topics from electronic health records}}
    \vskip 0.09in
    \vskip -5.5pt
    \rule{\textwidth}{1pt} \relax
    \vskip 0.05in
    \normalsize
    \begin{minipage}{0.9\textwidth}
    \begin{center}
        \textbf{Yixuan Li$^{1,2}$, Archer Y. Yang$^{1,2,4,*}$, Ariane Marelli$^{3,*}$, Yue Li$^{2,4,*}$}
    \end{center}
    $^1$Department of Mathematics and Statistics, McGill University, Montreal, Canada\\
    $^2$Mila - Quebec AI institute, Montreal, Canada\\
    $^3$McGill Adult Unit for Congenital Heart Disease (MAUDE Unit), McGill University of Health Centre, Montreal, Canada\\
    $^4$School of Computer Science, McGill University, Montreal, Canada\\
    $^*$Correspondence:\\ ariane.marelli@mcgill.ca, archer.yang@mcgill.ca, yueli@cs.mcgill.ca
    \end{minipage}
    \vskip 0.2in
}

\singlespacing

\section{Notation table}\label{Motation table}

\renewcommand{\arraystretch}{0.75} 
\small
\begin{tabular}{ll}
\hline 
\textbf{Notation}  & \textbf{Description} \tabularnewline
\hline 
$K$  & Total number of topics \tabularnewline
$M$  & Total number of EHR types \tabularnewline
$P$  & Total number of patients \tabularnewline
$m\in\{1,\ldots,M\}$  & Index for EHR types \tabularnewline
$V^{(m)}$  & Total number of unique EHR features for document type $m$ \tabularnewline
$k\in\{1,\ldots,K\}$  & Index for topics \tabularnewline
$j\in\{1,\ldots,P\}$  & Index for patient \tabularnewline
$N_{j}^{(m)}$  & Number of tokens in the EHR document of type $m$ for patient $j$ \tabularnewline
$i\in\{1,\ldots,N_{j}^{(m)}\}$  & Index for tokens for patient $j$ and document type $m$\tabularnewline
$\boldsymbol{\uppi}_{j}\in[0,1]^{K}$  & Phenotype prior for patient $j$ \tabularnewline
$\boldsymbol{\uptheta}_{j}\in[0,1]^{K}$  & Topic assignment for patient $j$ \tabularnewline
$\boldsymbol{\upalpha}\in\mathbb{R}_{+}^{K}$  & Hyperparameter for Dirichlet distribution of $\boldsymbol{\uptheta}_{j}$ \tabularnewline
$\phi_{kv}^{(m)}\in[0,1]$  & \makecell[l]{Feature distribution of token with index $v$ for topic $k$ and document \\
type $m$} \tabularnewline
$\boldsymbol{\upphi}_{k}^{(m)}\in[0,1]^{V^{(m)}}$  & Feature distribution for topic $k$ and document type $m$ \tabularnewline
$\boldsymbol{\upbeta}^{(m)}\in\mathbb{R}_{+}^{V^{(m)}}$  & Hyperparameter for Dirichlet distribution of $\boldsymbol{\upphi}_{k}^{(m)}$ \tabularnewline
$x_{ji}^{(m)}\in\{1,\ldots,V^{(m)}\}$  & Word index of token $i$ in the EHR document of type $m$ for patient
$j$ \tabularnewline
$z_{ji}^{(m)}\in\{1,\ldots,K\}$  & Latent topic assignment for token $i$ in document m for patient $j$ \tabularnewline
$\gamma_{jik}^{(m)}\in[0,1]$  & \makecell[l]{Variational probability of the $k^{th}$ topic assignment for token
$i$ of EHR \\ type $m$ for patient $j$} \tabularnewline
$\bar{\mf{z}}_{j}\in[0,1]^{K}$  & Average topic weight for patient $j$ \tabularnewline
$T_{j}\in\mathbb{R}_{+}$  & Observed time for patient $j$ \tabularnewline
$\delta_{j}\in\{0,1\}$  & Censoring status for patient $j$ \tabularnewline
$h_{0}\left(T_{j}\right)$  & Baseline hazard function for patient $j$\tabularnewline
$H_{0}\left(T_{j}\right)$  & Baseline cumulative hazard function for patient $j$\tabularnewline
$\mf{w}\in\mathbb{R}^{K}$  & Cox PH regression coefficient \tabularnewline
$\mathbf{T}\in\mathbb{R}_{+}^{P}$  & Vector of observed times for all patients \tabularnewline
$\boldsymbol{\updelta}\in\{0,1\}^{P}$  & Vector of censoring status for all patients \tabularnewline
$\mathcal{X}^{(m)}=\left\{ \left\{ x_{ji}^{(m)}\right\} _{i=1}^{N_{j}}\right\} _{j=1}^{P}$  & \makecell[l]{A set of $P$ lists of word indices for all tokens of EHR type $m$
for all \\ patients} \tabularnewline
$\mathcal{X}=\left\{ \mathcal{X}^{(m)}\right\} _{m=1}^{M}$  & The entire EHR data over the $M$ EHR types \tabularnewline
$\mathcal{Z}^{(m)}=\left\{ \left\{ z_{ji}^{(m)}\right\} _{i=1}^{N_{j}}\right\} _{j=1}^{P}$  & \makecell[l]{A set of $P$ lists of topic indices for all tokens of EHR type $m$
for all \\ patients} \tabularnewline
$\mathcal{Z}=\left\{ \mathcal{Z}^{(m)}\right\} _{m=1}^{M}$  & The topic assignments of the entire EHR data over the $M$ EHR types \tabularnewline
$\boldsymbol{\uppi}\in[0,1]^{P\times K}$  & Matrix of phenotype priors for all patients \tabularnewline
$\boldsymbol{\uptheta}\in[0,1]^{P\times K}$  & Matrix of topic assignments for all patients \tabularnewline
$\boldsymbol{\upphi}^{(m)}\in[0,1]^{K\times V^{(m)}}$  & Matrix of feature distributions for all topics of EHR type $m$ \tabularnewline
$\boldsymbol{\Phi}=\left\{ \boldsymbol{\upphi}^{(m)}\right\} _{m=1}^{M}$  & List of feature distribution over the $M$ EHR types\tabularnewline
$\mathcal{B}=\left\{ {\boldsymbol{\upbeta}^{(m)}}\right\} _{m=1}^{M}$  & List of hyperparameters for Dirichlet distribution of $\boldsymbol{\upphi}_{k}^{(m)}$ \tabularnewline
$\mf{U}\in\mathbb{R}^{P\times K}$  & Matrix of PheCode counts for all $P$ patients and $K$ PheCodes \tabularnewline
$u_{jk}$  & Count of the $k$-th PheCode for the $j$-th patient \tabularnewline
\hline 
\end{tabular}


\section{Generative process the model variants}\label{generating process}

\subsection{Generative process for MixEHR}

MixEHR follows the following generative process as illustrated in \fig~\ref{fig:MixEHRs_diagrams}a: 

\begin{enumerate}
    \item Generate patient-specific topic assignment $\boldsymbol{\uptheta}_{j}\sim\text{Dir} \left(\boldsymbol{\upalpha}\right), j=1,\ldots,P$
    \item Generate the feature distribution $\boldsymbol{\upphi}_{k}^{\left( m\right) }\sim \text{Dir}\left(\boldsymbol{\upbeta}^{\left( m\right) }\right)$ for topic $k=1,\ldots,K$ and type $m=1,\ldots, M$.
    \item For each of the EHR token $x_{ji}^{\left( m\right) }$, $i = 1, \ldots, N^{\left( m\right) }_{j}$ :
    \begin{enumerate}
        \item Generate a latent topic $z_{ji}^{\left( m\right) }\sim\text{Mul} \left(\boldsymbol{\uptheta}_{j}\right)$ 
        \item Generate a specific token $x_{ji}^{\left( m\right) }\sim\text{Mul}\left(\boldsymbol{\upphi}_{z_{ji}^{\left( m\right) }}^{\left( m\right) }\right)$
    \end{enumerate}
\end{enumerate}

\subsection*{Generative process for MixEHR-G}    

The generative process for MixEHR-G is illustrated in \fig~\ref{fig:MixEHRs_diagrams}b:

\begin{enumerate}
    \item Obtain the phenotype prior $\boldsymbol{\uppi}_{j}$ by a modified MAP \cite{liao2019high}
    algorithm
    \item Draw patient specific topic assignment $\boldsymbol{\uptheta}_{j}\sim\text{Dir}\left(\boldsymbol{\alpha} \odot \boldsymbol{\uppi}_{j}\right)$
    
    \item Generate the feature distribution $\boldsymbol{\upphi}_{k}^{\left( m\right) }\sim \text{Dir}\left(\boldsymbol{\upbeta}^{\left( m\right) }\right)$ for topic $k=1,\ldots,K$ and type $m=1,\ldots, M$.
    
    \item For each of the EHR token $x_{ji}^{ \left( m\right) }$, $i = 1, \ldots, N^{ \left( m\right) }_{j}$ :
    \begin{enumerate}
        \item Generate a latent topic $z_{ji}^{\left( m\right) }\sim\text{Mul}\left(\boldsymbol{\uptheta}_{j}\right)$ 
        \item Generate a specific token $x_{ji}^{\left( m\right) }\sim\text{Mul}\left(\boldsymbol{\upphi}_{z_{ji}^{\left( m\right) }}^{\left( m\right) }\right)$
    \end{enumerate}
\end{enumerate}

\subsection*{Generative process for MixEHR-Surv}

The generative process for MixEHR-Survival is illustrated in \fig~\ref{fig:MixEHRs_diagrams}c:

\begin{enumerate}
    \item Generate patient-specific topic assignment $\boldsymbol{\uptheta}_{j}\sim\text{Dir}\left(\boldsymbol{\alpha}\right)$
    \item Generate the feature distribution $\boldsymbol{\upphi}_{k}^{\left( m\right) }\sim \text{Dir}\left(\boldsymbol{\upbeta}^{\left( m\right) }\right)$ for topic $k=1,\ldots,K$ and type $m=1,\ldots, M$.
    \item For each of the EHR token $x_{ji}^{\left( m\right) }$, $i = 1, \ldots, N^{\left( m\right) }_{j}$ :
    \begin{enumerate}
        \item Generate a latent topic $z_{ji}^{\left( m\right) }\sim\text{Mul}\left(\boldsymbol{\uptheta}_{j}\right)$ 
        \item Generate a specific token $x_{ji}^{\left( m\right) }\sim\text{Mul}\left(\boldsymbol{\upphi}_{z_{ji}^{\left( m\right) }}^{\left( m\right) }\right)$
    \end{enumerate}
    \item Compute the average topic proportion for each patient: $\bar{\mf{z}}_j=[\bar{z}_{jk}]_{k=1}^{K}=\left[\frac{\sum_{m=1}^{M}\sum_{i=1}^{N_{j}^{(m)}}\mathbb{I}(z_{ji}^{\left( m\right) }=k)}{\sum_{m=1}^{M}N_{j}^{\left( m\right) }}\right]_{k=1}^K$
    \item Calculate the patient's hazard through the Cox proportional hazards
    model $h\left(T_{j}|\bar{\mf{z}}_j\right)=h_{0}\left(T_{j}\right)\exp\left\{ \mf{w}^{\top}\bar{\mf{z}}_j\right\}$,
    and we could further visualize the survival curve or estimate survival time using the median survival time.
\end{enumerate}

\subsection*{Generative process for \model}

The generative process for \model~ is illustrated in \fig~\ref{fig:MixEHRs_diagrams}d: 

\begin{enumerate}
    \item Obtain the phenotype prior $\boldsymbol{\uppi}_{j}$ by a modified MAP \cite{liao2019high}
    algorithm
    \item Draw patient specific topic assignment $\boldsymbol{\uptheta}_{j}\sim\text{Dir}\left(\boldsymbol{\alpha} \odot \boldsymbol{\uppi}_{j}\right)$
    \item Generate the feature distribution $\boldsymbol{\upphi}_{k}^{\left( m\right) }\sim \text{Dir}\left(\boldsymbol{\upbeta}^{\left( m\right) }\right)$ for topic $k=1,\ldots,K$ and type $m=1,\ldots, M$.
    
        \item For each of the EHR token $x_{ji}^{ \left( m\right) }$, $i = 1, \ldots, N^{ \left( m\right) }_{j}$ :
    \begin{enumerate}
        \item Generate a latent topic $z_{ji}^{\left( m\right) }\sim\text{Mul}\left(\boldsymbol{\uptheta}_{j}\right)$ 
        \item Generate a specific token $x_{ji}^{\left( m\right) }\sim\text{Mul}\left(\boldsymbol{\upphi}_{z_{ji}^{\left( m\right) }}^{\left( m\right) }\right)$
    \end{enumerate}
    \item Compute the average topic weight for each patient: 
    $$\bar{\mf{z}}_j=[\bar{z}_{jk}]_{k=1}^{K}=\left[\frac{\sum_{m=1}^{M}\sum_{i=1}^{N_{j}^{(m)}}\mathbb{I}(z_{ji}^{\left( m\right) }=k)}{\sum_{m=1}^{M}N_{j}^{\left( m\right) }}\right]_{k=1}^K$$
    \item Calculate the patient's hazard through the Cox proportional hazards model $h\left(T_{j}|\bar{\mf{z}}_{j}\right)=h_{0}\left(T_{j}\right)\exp\left\{ \mf{w}^{\top}\bar{\mf{z}}_{j}\right\} $,
     we could further visualize the survival curve or estimate survival time using the median survival time.
\end{enumerate}

\section{Computing PheCode topic priors}\label{sec:prior}


We compute $\pi_{jk} = p(y_{jk} = 1 \mid u_{jk})$ for each patient $j$ and topic $k$ in 3 steps:
\begin{itemize}
    \item Step 1: After mapping each ICD code to its corresponding PheCode (\url{https://phewascatalog.org/phecodes}), we calculate the PheCode counts $u_{jk}$ for each patient, denoted by $j$, where $j = 1, \ldots, P$, across each PheCode, denoted by $k$, where $k = 1, \ldots, K$. It's important to note that for a patient who encounters the same PheCode multiple times, either due to repeated ICD code mappings or multiple healthcare visits, each instance is individually accounted for. This approach results in the possibility of accruing multiple counts for the same PheCode for a single patient. As a result, we convert the $P\times V^{(\text{ICD})}$ to a $P \times K$ matrix $\boldsymbol{U} = [u_{jk}]_{P \times K}$. We then infer the posterior distribution of $y_{jk}$ in two parallel ways.
   
    \item Step 2A (Model A): Assuming that the counts for a PhenoCode $k$ follows a Poisson distribution with parameters $\pi_{jk}$, $\rho_0$ and $\rho_1$. The Poisson likelihood takes the following form:
    \begin{align}    
        P\left(u_{jk}\right)
        =\pi_{jk}\frac{\left(\rho_1\right)^{u_{jk}}e^{-\rho_1}}{u_{jk}!}+\left(1-\pi_{jk}\right)\frac{\left(\rho_0\right)^{u_{jk}}e^{-\rho_0}}{u_{jk}!}, 
    \end{align}
where $\pi_{jk}$ corresponds to the foreground Poisson component with larger mean $\rho_1$ and and $1-\pi_{jk}$ corresponds to the population background Poisson with lower mean $\rho_0$. Given data $\{u_{jk}\}_{j=1}^P$, we perform expectation-maximization (EM) algorithm: in the E-step, we infer the posterior probability $\widehat{\pi}_{jk}=\hat{p}(y_{jk}=1 | u_{jk})$ and in the M-step, we maximize the likelihood with respect to $\rho_1$ and $\rho_0$.
\item Step 2B (Model B): 
Alternatively, we can assume that for each PheCode $k$, the log-transformed count data $g(u_{1k}),\ldots,g(u_{Pk})$, with $g(u) = \log(u) + 1$ follows a two-component univariate Gaussian mixture model:
\begin{align}
    P\left(g(u_{jk}) = x\right)=\frac{\pi'_{jk}}{\sqrt{2\pi\sigma_{1}^{2}}}\exp\left(-\frac{\left(x-\mu_{1}\right)^{2}}{2\sigma_{1}^{2}}\right) 
    + 
    \frac{1 - \pi'_{jk}}{\sqrt{2\pi\sigma_{0}^{2}}}\exp\left(-\frac{\left(x-\mu_{0}\right)^{2}}{2\sigma_{0}^{2}}\right)
\end{align}
We then perform EM algorithm to alternate between inferring $\widehat{\pi}'_{jk}=\hat{p}(y'_{jk}=1|u_{jk})$ and computing maximum likelihood estimates for the Gaussian parameters.

\item Step 3: The prior probability for a patient $j$ having phenotype
 $k$ is set to $\pi_{jk}=\frac{1}{2}\left(\widehat{\pi}_{jk}+ \widehat{\pi}'_{jk}\right)$. 
\end{itemize}

In the application of the MIMIC-III data, as it is not a longitudinal dataset, each PheCode was documented no more than once for each patient. In this case, we assigned the hyperparameters $\pi_{jk}$ for each phenotype $k$ as either one or zero, based on whether the corresponding PheCode was observed or not for patient $j$, respectively.

\section{Details of stochastic joint collapsed variational Bayesian inference}\label{sgd}

First, we derive the  joint-likelihood function of all the
parameters for observational data and latent variables conditioned on
priors and survival regression coefficients for \model~ (Fig \ref{fig:MixEHRs_diagrams}d)
model:
\begin{align*}
& \qquad p\left(\mf{T},\boldsymbol{\updelta},\mathcal{X}, \mathcal{Z}, \boldsymbol{\uptheta}, \boldsymbol{\Phi}\mid\boldsymbol{\alpha},\boldsymbol{\uppi},\mathcal{B},h_{0}(\cdot),\mf{w}\right)\\
& =  \underset{\text{supervised part}}{\underbrace{p\left(\mf{T},\boldsymbol{\updelta}\mid\mathcal{Z},h_{0}(\cdot),\mf{w}\right)}}\underset{\text{unsupervised part}}{\underbrace{p\left(\mathcal{X}, \mathcal{Z}, \boldsymbol{\uptheta}, \boldsymbol{\Phi}\mid\boldsymbol{\alpha},\boldsymbol{\uppi},\mathcal{B}\right)}}
\end{align*}
where for the survival supervised part, we use the Cox proportional      
hazards (PH) model with elastic net penalization for the survival
coefficients. The full likelihood function of the penalized Cox PH
model is obtained by incorporating Breslow's estimate of the baseline
hazard function.
\begin{align*}
&    p\left(\mf{T},\boldsymbol{\updelta}\mid \mathcal{Z},h_{0}(\cdot), \mf{w} \right) \\
= & \prod_{j=1}^P p\left( T_j, \delta_j \mid \bar{\mf{z}}_j, h_0(T_j), \mf{w}\right) \\
= & \prod_{j=1}^P \left[ h\left( T_j, \bar{\mf{z}}_j\right)\right]^{\delta_j} S\left(T_j,\bar{\mf{z}}_j\right)\exp\left\{-\lambda_2\|\mf{w}\|^2_2 - \lambda_1\|\mf{w}\|_1\right\}\\
= & \prod_{j=1}^P\left\{ \left[h_{0}\left(T_{j}\right)\exp\left(\mf{w}^{\top}\bar{\mf{z}}_j\right)\right]^{\delta_{j}}\times\exp\left[-H_{0}\left(T_{j}\right)\exp\left(\mf{w}^{\top}\bar{\mf{z}}_j\right)\right]\right\} \exp\left\{-\lambda_2\|\mf{w}\|^2_2 - \lambda_1\|\mf{w}\|_1\right\}.
\end{align*}
Here $H_{0}\left(t\right)$ denotes the cumulative baseline hazard
function, obtained by the integral of the baseline hazard function
between integration limits of $0$ and $t$ as $H_{0}\left(t\right)=\int_{0}^{t}h_{0}\left(u\right)du$. The elastic net penalty terms including $\|\mf{w}\|^2_2=\sum_kw^2_k$ and $\|\mf{w}\|_1=\sum_k |w_k|$ consist of the L2 and L1 regularization term weighted by the hyperparameters $\lambda_2$ and $\lambda_1$, respectively.

We will use the collapsed variational inference algorithm to integret
out $\boldsymbol{\uptheta}$ and $\boldsymbol{\Phi}$ in the joint likelihood function to achieve more
accurate and efficient inference \cite{griffiths2004finding}. This
is due to the conjugacy of Dirichlet variables $\boldsymbol{\uptheta}$
and $\boldsymbol{\Phi}$ to the multinomial likelihood variables $\mathcal{X}$
and $\mathcal{Z}$. 
\begin{align*}
 & p\left(\mf{T},\boldsymbol{\updelta},\mathcal{X},\mathcal{Z}\mid\boldsymbol{\alpha},\boldsymbol{\uppi},\mathcal{B},h_{0}(\cdot),\mf{w}\right)\\
= & p\left(\mf{T},\boldsymbol{\updelta}\mid\mathcal{Z},h_{0}(\cdot),\mf{w}\right)p\left(\mathcal{X},\mathcal{Z}\mid\boldsymbol{\alpha},\boldsymbol{\uppi},\mathcal{B}\right)\\
= & p\left(\mf{T},\boldsymbol{\updelta}\mid\mathcal{Z},h_{0}(\cdot),\mf{w}\right)\int\int p\left(\mathcal{X}, \mathcal{Z}, \boldsymbol{\uptheta}, \boldsymbol{\Phi}\mid\boldsymbol{\alpha},\boldsymbol{\uppi},\mathcal{B}\right)d\boldsymbol{\Phi}d\boldsymbol{\uptheta}\\
= & p\left(\mf{T},\boldsymbol{\updelta}\mid\mathcal{Z},h_{0}(\cdot),\mf{w}\right)\int\int p\left(\mathcal{X}\mid\mathcal{Z},\boldsymbol{\Phi}\right)p\left(\boldsymbol{\Phi}\mid\mathcal{B}\right)p\left(\mathcal{Z}\mid\boldsymbol{\uptheta}\right)p\left(\boldsymbol{\uptheta}\mid\boldsymbol{\alpha},\boldsymbol{\uppi}\right)d\boldsymbol{\Phi}d\boldsymbol{\uptheta}\\
= & p\left(\mf{T},\boldsymbol{\updelta}\mid\mathcal{Z},h_{0}(\cdot),\mf{w}\right)\int p\left(\mathcal{X}\mid\mathcal{Z},\boldsymbol{\Phi}\right)p\left(\boldsymbol{\Phi}\mid\mathcal{B}\right)d\boldsymbol{\Phi}\times\int p\left(\mathcal{Z}\mid\boldsymbol{\uptheta}\right)p\left(\boldsymbol{\uptheta}\mid\boldsymbol{\alpha},\boldsymbol{\uppi}\right)d\boldsymbol{\uptheta}
\end{align*}

Upon substituting the distributions outlined in the generative process of \model, as detailed in \sect~\ref{generating process}, the integral can be evaluated as follows:

\begin{align*}
    & \int p\left(\mathcal{Z}\mid\boldsymbol{\uptheta}\right)p\left(\boldsymbol{\uptheta}\mid\boldsymbol{\alpha},\boldsymbol{\uppi}\right)d\boldsymbol{\uptheta} \\
    = & \int \left(\prod_{j=1}^P \prod_{k=1}^K \theta_{j k}^{n_{j \bullet k}^{(\bullet)}} \right) \times \left(\prod_{j=1}^P \frac{\Gamma\left(\sum_{k=1}^K\alpha_{k}\boldsymbol{\uppi}_j\right)}{\prod_{k=1}^K\Gamma\left(\alpha_{k}\boldsymbol{\uppi}_j\right)} \prod_{k=1}^K \theta_{j k}^{\alpha_k \pi_{j k}-1} \right) d\boldsymbol{\uptheta} \\
    = & \prod_{j=1}^P  \frac{\Gamma\left(\sum_{k=1}^K\alpha_{k}\boldsymbol{\uppi}_j\right)}{\prod_{k=1}^K\Gamma\left(\alpha_{k}\boldsymbol{\uppi}_j\right)} \int \left(\prod_{k=1}^K \theta_{j k}^{\alpha_k \pi_{j k}-1+n_{j \bullet k}^{(\bullet)}} \right)d\boldsymbol{\uptheta} \\
    = & \prod_{j=1}^P \frac{\Gamma\left(\sum_{k=1}^K\alpha_{k}\boldsymbol{\uppi}_j\right)}{\prod_{k=1}^K\Gamma\left(\alpha_{k}\boldsymbol{\uppi}_j\right)}\frac{\prod_{k=1}^K\Gamma\left(\alpha_{k}\boldsymbol{\uppi}_j+n_{j \bullet k}^{(\bullet)}\right)}{\Gamma\left(\sum_{k=1}^K\boldsymbol{\alpha}_{k}\boldsymbol{\uppi}_j+n_{j \bullet k}^{(\bullet)}\right)}
\end{align*}

\begin{align*}
    & \int p\left(\mathcal{X}\mid\mathcal{Z},\boldsymbol{\Phi}\right)p\left(\boldsymbol{\Phi}\mid\mathcal{B}\right)d\boldsymbol{\Phi} \\
    & = \int \left(\prod_{m=1}^M\prod_{k=1}^K\prod_{v=1}^{V^{(m)}} \phi_{vk}^{(m)n_{\bullet vk}^{(m)}}\right) \times \left(\prod_{m=1}^M\prod_{k=1}^K \frac{\Gamma\left(\sum_{v=1}^{V^{(m)}}\beta_{v}^{(m)}\right)}{\prod_{v=1}^{V^{(m)}}\Gamma\left(\beta_{v}^{(m)}\right)} \prod_{v=1}^{V^{(m)}} \phi_{vk}^{(m)\beta_{v}^{(m)}-1}\right) d\boldsymbol{\Phi} \\
    & = \prod_{m=1}^M\prod_{k=1}^K\frac{\Gamma\left(\sum_{v=1}^{V^{(m)}}\beta_{v}^{(m)}\right)}{\prod_{v=1}^{V^{(m)}}\Gamma\left(\beta_{v}^{(m)}\right)} \int \left(\prod_{v=1}^{V^{(m)}} \phi_{vk}^{(m)\beta_{v}^{(m)}-1+n_{\bullet vk}^{(m)}}\right)d\boldsymbol{\Phi} \\
    & = \prod_{k=1}^K\prod_{m=1}^M\frac{\Gamma\left(\sum_{v=1}^{V^{(m)}}\beta_{v}^{\left(m\right)}\right)}{\prod_{v=1}^{V}\Gamma\left(\beta_{v}^{\left(m\right)}\right)}\frac{\prod_{v=1}^{V^{(m)}}\Gamma\left(\beta_{v}^{\left(m\right)}+n_{\bullet vk}^{(m)}\right)}{\Gamma\left(\sum_{v=1}^{V^{(m)}} \beta_{v}^{\left(m\right)}+n_{\bullet vk}^{(m)}\right)}
\end{align*}

where the coordinate sufficient statistics are:
$$
    n_{\bullet vk}^{(m)}=\sum_{j=1}^{P}\sum_{i=1}^{N_{j}^{(m)}}\mathbb{I}\left[x_{ji}^{(m)}=v,z_{ji}^{(m)}=k\right]
$$
$$
    n_{j \bullet k}^{(\bullet)}=\sum_{m=1}^{M}\sum_{i=1}^{N_{j}^{(m)}}\mathbb{I}\left[z_{ji}^{(m)}=k\right]
$$

Thus, we have:
\begin{align*}
    &p\left(\mathcal{X},\mathcal{Z}\mid\boldsymbol{\alpha},\boldsymbol{\uppi},\mathcal{B}\right)\\
    = & \prod_{k=1}^K\prod_{m=1}^M\frac{\Gamma\left(\sum_{v=1}^{V^{(m)}}\beta_{v}^{\left(m\right)}\right)}{\prod_{v=1}^{V^{(m)}}\Gamma\left(\beta_{v}^{\left(m\right)}\right)}\frac{\prod_{v=1}^{V^{(m)}}\Gamma\left(\beta_{v}^{\left(m\right)}+n_{\bullet vk}^{(m)}\right)}{\Gamma\left(\sum_{v=1}^{V^{(m)}} \beta_{v}^{\left(m\right)}+n_{\bullet vk}^{(m)}\right)} \prod_{j=1}^P \frac{\Gamma\left(\sum_{k=1}^K\alpha_{k}\boldsymbol{\uppi}_j\right)}{\prod_{k=1}^K\Gamma\left(\alpha_{k}\boldsymbol{\uppi}_j\right)}\frac{\prod_{k=1}^K\Gamma\left(\alpha_{k}\boldsymbol{\uppi}_j+n_{j \bullet k}^{(\bullet)}\right)}{\Gamma\left(\sum_{k=1}^K\boldsymbol{\alpha}_{k}\boldsymbol{\uppi}_j+n_{j \bullet k}^{(\bullet)}\right)}
\end{align*}

Then, we will derive the evidence lower bound (ELBO) for the current
marginal distribution for the observational data as follows:


\begin{align*}
    \mathcal{L}_{ELBO} & \equiv  \mathbb{E}_{q(\mathcal{Z})}\log p\left(\mf{T},\boldsymbol{\updelta},\mathcal{X},\mathcal{Z}\mid\boldsymbol{\alpha},\boldsymbol{\uppi},\mathcal{B},h_{0}(\cdot),\mf{w}\right)-\mathbb{E}_{q(\mathcal{Z})}\log q\left(\mathcal{Z}\right) \\
     & = \underset{\mathcal{Z}}{\sum}q\left(\mathcal{Z}\right)\log p\left(\mf{T},\boldsymbol{\updelta},\mathcal{X},\mathcal{Z}\mid\boldsymbol{\alpha},\boldsymbol{\uppi},\mathcal{B},h_{0}(\cdot),\mf{w}\right)-\underset{\mathcal{Z}}{\sum}q\left(\mathcal{Z}\right)\log q\left(\mathcal{Z}\right)
\end{align*}
Maximizing $\mathcal{L}_{ELBO}$ is equivalent to minimizing
the Kullback--Leibler (KL) divergence, as they sum up as the joint
distribution of the observational data which is a constant:
\begin{align*}
    \mathcal{K}\mathcal{L}[q(\mathcal{Z})\|p\left(\mf{T},\boldsymbol{\updelta},\mathcal{X},\mathcal{Z}\right)] & =\mathbb{E}_{q(\mathcal{Z})}\log q\left(\mathcal{Z}\right)-\mathbb{E}_{(\mathcal{Z})}\log p\left(\mf{T},\boldsymbol{\updelta},\mathcal{X},\mathcal{Z}\mid\boldsymbol{\alpha},\boldsymbol{\uppi},\mathcal{B},h_{0}(\cdot),\mf{w}\right)+\log p\left(\mf{T},\boldsymbol{\updelta},\mathcal{X}\right)\\
     & =-\mathcal{L}_{ELBO}+\log p\left(\mf{T},\boldsymbol{\updelta},\mathcal{X}\right)
\end{align*}
The mean-field assumption pertains only to word-specific topic assignments $\mathcal{Z}$, which have the proposed distribution under the variational parameter $\gamma_{jik}^{(m)}$ as defined below:
\begin{align*}
    q(\mathcal{Z}) = \prod_{m=1}^M\prod_{j=1}^{P}\prod_{i=1}^{N_j^{(m)}} q(z_{ji}^{(m)}\mid \gamma_{ji1}^{(m)},\ldots,\gamma_{jiK}^{(m)}) = 
    \prod_{m=1}^M\prod_{j=1}^{P}\prod_{i=1}^{N_j^{(m)}} \prod_{k=1} ^K\gamma_{jik} ^{{(m)}^{\mathbb{I}[z_{ji}^{(m)}=k]}}
\end{align*}

Under the mean-field assumption, maximizing the ELBO with respect to $\gamma_{jik}^{\left(m\right)}$ is equivalent to calculating the variational expectation $\mathbb{E}_{q(\mathcal{Z})}[z_{ji}^{\left(m\right)}=k]$ conditioned on the variational expected value for other tokens \cite{teh2006collapsed,sato2012rethinking}. The coordinate ascent update has an approximate closed-form expression as derived below:
\begin{align*}
\gamma_{jik}^{\left(m\right)} &= \frac{\exp \left\{\mathbb{E}_{q\left(z_{(j,-i)}^{\left(m\right)}\right)}\left[ \log  p\left(\mf{T},\boldsymbol{\updelta},\mathcal{X},\mathcal{Z}\mid\boldsymbol{\alpha},\boldsymbol{\uppi},\mathcal{B},h_{0}(\cdot),\mf{w}\right)\right] \right\}}{\exp \left\{\int \mathbb{E}_{q\left(z_{(j,-i)}^{\left(m\right)}\right)}\left[\log p\left(\mf{T},\boldsymbol{\updelta},\mathcal{X},\mathcal{Z}\mid\boldsymbol{\alpha},\boldsymbol{\uppi},\mathcal{B},h_{0}(\cdot),\mf{w}\right)\right] d z_{ji}^{\left(m\right)}\right\}} \\
&\propto \exp \left\{\mathbb{E}_{q\left(z_{(j,-i)}^{\left(m\right)}\right)}\left[ \log p\left(\mf{T},\boldsymbol{\updelta},\mathcal{X},\mathcal{Z}\mid\boldsymbol{\alpha},\boldsymbol{\uppi},\mathcal{B},h_{0}(\cdot),\mf{w}\right)\right]\right\}
\end{align*}

Then we aximizing the ELBO with respect to $\gamma_{jik}^{\left(m\right)}$, 
\begin{align*}
\log\gamma_{jik}^{\left(m\right)} & =\mathbb{E}_{q\left(z_{(j,-i)}^{\left(m\right)}\right)}\left[\log p\left(\mf{T},\boldsymbol{\updelta},\mathcal{X},\mathcal{Z}\mid\boldsymbol{\alpha},\boldsymbol{\uppi},\mathcal{B},h_{0}(\cdot),\mf{w}\right)\right]+\text{const}\\
 & =\mathbb{E}_{q\left(z_{(j,-i)}^{\left(m\right)}\right)}\left[\log\left(p\left(\mf{T},\boldsymbol{\updelta}\mid\mathcal{Z},h_{0}(\cdot),\mf{w}\right)p\left(\mathcal{X},\mathcal{Z}\mid\boldsymbol{\alpha},\boldsymbol{\uppi},\mathcal{B}\right)\right)\right]+\text{const}\\
 & =\mathbb{E}_{q\left(z_{(j,-i)}^{\left(m\right)}\right)}\left[\log p\left(T_{j},\delta_{j}\mid z_{(j,-i)}^{\left(m\right)},z_{ji}^{\left(m\right)}=k,h_{0}(\cdot),\mf{w}\right)\right]\\
 & \qquad+\mathbb{E}_{q\left(z_{(j,-i)}^{\left(m\right)}\right)}\left[\log p\left(\mathcal{X},\mathcal{Z}\mid\boldsymbol{\alpha},\boldsymbol{\uppi},\mathcal{B}\right)\right]+\text{const}\\
 & =\mathbb{E}_{q\left(z_{(j,-i)}^{\left(m\right)}\right)}\left[\log p\left(T_{j},\delta_{j}\mid z_{(j,-i)}^{\left(m\right)},z_{ji}^{\left(m\right)}=k,h_{0}(\cdot),\mf{w}\right)\right]\\
 & +\mathbb{E}_{q\left(z_{(j,-i)}^{\left(m\right)}\right)}\Bigg[\log\Bigg(\prod_{k=1}^{K}\prod_{m=1}^{M}\frac{\Gamma\left(\sum_{v=1}^{V^{(m)}}\beta_{v}^{\left(m\right)}\right)}{\prod_{v=1}^{V^{(m)}}\Gamma\left(\beta_{v}^{\left(m\right)}\right)}\frac{\prod_{v=1}^{V^{(m)}}\Gamma\left(\beta_{v}^{\left(m\right)}+n_{\bullet vk}^{(m)}\right)}{\Gamma\left(\sum_{v=1}^{V^{(m)}}\beta_{v}^{(m)}+n_{\bullet vk}^{(m)}\right)}\\
 & \quad\prod_{j=1}^{P}\frac{\Gamma\left(\sum_{k}\alpha_{k}\boldsymbol{\uppi}_{j}\right)}{\prod_{k=1}^{K}\Gamma\left(\alpha_{k}\boldsymbol{\uppi}_{j}\right)}\frac{\prod_{k=1}^{K}\Gamma\left(\alpha_{k}\boldsymbol{\uppi}_{j}+n_{j\bullet k}^{(\bullet)}\right)}{\Gamma\left(\sum_{k=1}^{K}\alpha_{k}\boldsymbol{\uppi}_{j}+n_{j\bullet k}^{(\bullet)}\right)}\Bigg)\Bigg]+\text{const}
\end{align*}

Thus, we calculate the expontential spontaneously at both side
\begin{align*}
\gamma_{jik}^{\left(m\right)} & \propto\exp\left\{ \mathbb{E}_{q\left(z_{(j,-i)}^{\left(m\right)}\right)}\left[\log p\left(T_{j},\delta_{j}\mid z_{(j,-i)}^{\left(m\right)},z_{ji}^{\left(m\right)}=k,h_{0}(\cdot),\mf{w}\right)\right]\right\} \\
 & \qquad\exp\Bigg\{\mathbb{E}_{q\left(z_{(j,-i)}^{\left(m\right)}\right)}\Bigg[\log\Bigg(\prod_{k=1}^{K}\prod_{m=1}^{M}\frac{\Gamma\left(\sum_{v=1}^{V^{(m)}}\beta_{v}^{\left(m\right)}\right)}{\prod_{v=1}^{V^{(m)}}\Gamma\left(\beta_{v}^{\left(m\right)}\right)}\frac{\prod_{v=1}^{V^{(m)}}\Gamma\left(\beta_{v}^{\left(m\right)}+n_{\bullet vk}^{(m)}\right)}{\Gamma\left(\sum_{v=1}^{V^{(m)}}\beta_{v}^{(m)}+n_{\bullet vk}^{(m)}\right)}\\
 & \qquad\prod_{j=1}^{P}\frac{\Gamma\left(\sum_{k}\alpha_{k}\boldsymbol{\uppi}_{j}\right)}{\prod_{k=1}^{K}\Gamma\left(\alpha_{k}\boldsymbol{\uppi}_{j}\right)}\frac{\prod_{k=1}^{K}\Gamma\left(\alpha_{k}\boldsymbol{\uppi}_{j}+n_{j\bullet k}^{(\bullet)}\right)}{\Gamma\left(\sum_{k=1}^{K}\alpha_{k}\boldsymbol{\uppi}_{j}+n_{j\bullet k}^{(\bullet)}\right)}\Bigg)\Bigg]\Bigg\}
\end{align*}
where the footnote $\left(j,-i\right)$ denote when we calculating
the coordinate sufficient statistics, we exclude the variable with
index $ji$. 

We choose the survival model as the Cox proportional hazards
model. The corresponding hazard function and survival function
could be written as
$$
h\left(T_{j},\bar{\mf{z}}_j\right)=h_{0}\left(T_{j}\right)\exp\left(\mf{w}^{\top}\bar{\mf{z}}_j\right)
$$
and 
$$
S\left(T_{j},\bar{\mf{z}}_j\right)=\exp\left[-H_{0}\left(T_{j}\right)\exp\left(\mf{w}^{\top}\bar{\mf{z}}_j\right)\right]
$$
respectively. The vector $\mf{w}\in\mathbb{R}^K$ contains the survival coefficients, and  $h_{0}\left(T_{j}\right)$ is the baseline hazard at time $T_j$. $H_{0}\left(T_{j}\right)$ denotes the cumulative hazard at time $T_j$, which is obtained by the integral of the baseline hazard function between integration limits of $0$ and $t$ as $H_{0}\left(t\right)=\int_{0}^{t}h_{0}\left(u\right)du$. Under those settings, we could further derive the supervised part
as follows:
\begin{align*}
     & \qquad\mathbb{E}_{q\left(z_{(j,-i)}^{\left(m\right)}\right)}\left[\log p\left(T_j ,\delta_j \mid z_{(j,-i)}^{\left(m\right)},z_{ji}^{\left(m\right)}=k,h_{0}(\cdot),\mf{w}\right)\right]\\
     & \overset{(i)}{=} \mathbb{E}_{q\left(z_{(j,-i)}^{\left(m\right)}\right)}\left[ \log p\left(T_j ,\delta_j \mid \bar{\mf{z}}_{(j,-i)}^{\left(m\right)},\bar{\mf{z}}_{ji}^{\left(m\right)}, h_{0}(\cdot),\mf{w}\right) \right]\\
     & = \mathbb{E}_{q\left(z_{(j,-i)}^{\left(m\right)}\right)} \left[ \log\left(h\left(T_j , \bar{\mf{z}}_{(j,-i)}^{\left(m\right)},\bar{\mf{z}}_{ji}^{\left(m\right)}\right)^{\delta_j} S\left(T_j, \bar{\mf{z}}_{(j,-i)}^{\left(m\right)},\bar{\mf{z}}_{ji}^{\left(m\right)}\right)\right) \right]\\ 
     & = \mathbb{E}_{q\left(z_{(j,-i)}^{\left(m\right)}\right)} \left[ \delta_{j}\log h_{0}\left(T_{j}\right) + \delta_{j}\mf{w}^{\top}\bar{\mf{z}}_{(j,-i)}^{\left(m\right)} 
     + \delta_{j}\mf{w}^{\top}\bar{\mf{z}}_{ji}^{\left(m\right)}- H_0\left(T_j\right)\exp\left(\mf{w}^{\top}\left( \bar{\mf{z}}_{(j,-i)}^{\left(m\right)} + \bar{\mf{z}}_{ji}^{\left(m\right)}\right)\right)\right] \\
     & \overset{(ii)}{=} \delta_{j}\log h_{0}\left(T_{j}\right)+\delta_{j}\mathbb{E}_{q\left(z_{(j,-i)}^{\left(m\right)}\right)}\left[\mf{w}^{\top}\bar{\mf{z}}_{(j,-i)}^{\left(m\right)}\right]+\delta_{j}\frac{w_{k}}{N_{j}^{\left(m\right)}}-H_{0}\left(T_{j}\right)\mathbb{E}_{q\left(z_{(j,-i)}^{\left(m\right)}\right)}\left[\exp\left(\mf{w}^{\top}\bar{\mf{z}}_{(j,-i)}^{\left(m\right)}+\frac{w_{k}}{N_{j}^{\left(m\right)}}\right)\right]  \\
     &= \delta_{j}\mathbb{E}_{q\left(z_{(j,-i)}^{\left(m\right)}\right)}\left[\mf{w}^{\top}\bar{\mf{z}}_{(j,-i)}^{\left(m\right)}\right] +\delta_{j}\frac{w_{k}}{N_{j}^{\left(m\right)}}-H_{0}\left(T_{j}\right)\mathbb{E}_{q\left(z_{(j,-i)}^{\left(m\right)}\right)}\left[\exp\left(\mf{w}^{\top}\bar{\mf{z}}_{(j,-i)}^{\left(m\right)}\right)\right]\exp\left(\frac{w_{k}}{N_{j}^{\left(m\right)}}\right)  + \text{const}  \\
     &  \overset{(iii)}{\approx} \delta_{j}\mathbb{E}_{q\left(z_{(j,-i)}^{\left(m\right)}\right)}\left[\mf{w}^{\top}\bar{\mf{z}}_{(j,-i)}^{\left(m\right)}\right] + \delta_{j}\frac{w_{k}}{N_{j}^{\left(m\right)}}-H_{0}\left(T_{j}\right)\mathbb{E}_{q\left(z_{(j,-i)}^{\left(m\right)}\right)}\left[\mf{w}^{\top}\bar{\mf{z}}_{(j,-i)}^{\left(m\right)}+1\right]\exp\left(\frac{w_{k}}{N_{j}^{\left(m\right)}}\right)  + \text{const} \\
     & \approx \delta_{j}\mathbb{E}_{q\left(z_{(j,-i)}^{\left(m\right)}\right)}\left[\mf{w}^{\top}\bar{\mf{z}}_{j}^{\left(m\right)}\right] + \delta_{j}\frac{w_{k}}{N_{j}^{\left(m\right)}}-H_{0}\left(T_{j}\right)\mathbb{E}_{q\left(z_{(j,-i)}^{\left(m\right)}\right)}\left[\mf{w}^{\top}\bar{\mf{z}}_{j}^{\left(m\right)}+1\right]\exp\left(\frac{w_{k}}{N_{j}^{\left(m\right)}}\right)  + \text{const} \\
     & \overset{(iv)}{=} \delta_j\mf{w}^{\top}\bar{\boldsymbol{\gamma}}_{j}^{\left(m\right)}+ \delta_{j}\frac{w_{k}}{N_{j}^{\left(m\right)}} -H_{0}\left(T_{j}\right)\left(\mf{w}^{\top}\bar{\boldsymbol{\gamma}}_{j}^{\left(m\right)}+1\right)\exp\left(\frac{w_{k}}{N_{j}^{\left(m\right)}}\right)  + \text{const} 
\end{align*}

The equation $(i)$ follows by defining \[\bar{\mf{z}}_{j i}^{(m)}=\left[\frac{\mathbb{I}\left(z_{j i}^{(m)}=k\right)}{N_j^{(m)}}\right]_{k=1}^K,\]
and 
\[\bar{\mf{z}}_{(j,-i)}^{(m)}=\left[\frac{\sum_{i'=1}^{N_{j}^{(m)}}\mathbb{I}\left((z_{j i'}^{(m)}=k) \cap (i' \neq i)\right)}{N_j^{(m)}}\right]_{k=1}^K.\]

The equation $(ii)$ follows by 
\[\left[ \bar{\mf{z}}_{j i}^{(m)} \right]_k = \frac{\mathbb{I}\left(z_{j i}^{(m)}=k\right)}{N_j^{(m)}} =\frac{1}{N_j^{(m)}}\] 
and 
\[\left[ \bar{\mf{z}}_{j i}^{(m)} \right]_{k'} = \frac{\mathbb{I}\left(z_{j i}^{(m)}=k'\right)}{N_j^{(m)}} =0,\]
for $k' \neq k$, since $z_{ji}^{(m)} = k$.

The approximation $(iii)$ is due to the first-order Taylor series of the exponential term $\exp\left(\mf{w}^{\top}\left[\bar{\mf{z}}_{(j,-i)}^{\left(m\right)}\right]_{j}\right)$. Note that the exponential function can be approximated by Taylor series as 
$\exp(x)=1 + x + x^2/2! + x^2/3! + \ldots$. For computational efficiency, we only took the first order of the Taylor series, which correspond to the first two terms $1+x$.

The equation $(iv)$ follows by defining: 

\begin{align*}
\bar{\boldsymbol{\gamma}}_{j}^{\left(m\right)} & =[\bar{\gamma}_{jk}^{(m)}]_{k=1}^{K}=\left[\frac{\sum_{i=1}^{N_{j}^{(m)}}\gamma_{jik}^{\left(m\right)}}{N_{j}^{\left(m\right)}}\right]_{k=1}^{K}\\
 & =\left[\frac{\sum_{i=1}^{N_{j}^{(m)}}\mathbb{E}_{q\left(z_{(j,-i)}^{\left(m\right)}\right)}\left[\mathbb{I}\left(z_{ji}^{(m)}=k\right)\right]}{N_{j}^{\left(m\right)}}\right]_{k=1}^{K}\\
 & =\mathbb{E}_{q\left(z_{(j,-i)}^{\left(m\right)}\right)}\left[\bar{\mf{z}}_{j}^{\left(m\right)}\right]
\end{align*}

And the expectation of the unsupervised part could be derived as:
\begin{align*}
 & \qquad\mathbb{E}_{q\left(z_{(j,-i)}^{(m)}\right)}\Bigg[\log\Bigg(\prod_{k=1}^{K}\prod_{m=1}^{M}\frac{\Gamma\left(\sum_{v=1}^{V^{(m)}}\beta_{v}^{(m)}\right)}{\prod_{v=1}^{V^{(m)}}\Gamma\left(\beta_{v}^{(m)}\right)}\frac{\prod_{v=1}^{V^{(m)}}\Gamma\left(\beta_{v}^{(m)}+n_{\bullet vk}^{(m)}\right)}{\Gamma\left(\sum_{v=1}^{V^{(m)}}\beta_{v}^{(m)}+n_{\bullet vk}^{(m)}\right)}\\
 & \quad\times\prod_{j=1}^{P}\frac{\Gamma\left(\sum_{k}\alpha_{k}\boldsymbol{\uppi}_{j}\right)}{\prod_{k=1}^{K}\Gamma\left(\alpha_{k}\boldsymbol{\uppi}_{j}\right)}\frac{\prod_{k=1}^{K}\Gamma\left(\alpha_{k}\boldsymbol{\uppi}_{j}+n_{j\bullet k}^{(\bullet)}\right)}{\Gamma\left(\sum_{k=1}^{K}\alpha_{k}\boldsymbol{\uppi}_{j}+n_{j\bullet k}^{(\bullet)}\right)}\Bigg)\Bigg]\\
 & =\mathbb{E}_{q\left(z_{(j,-i)}^{(m)}\right)}\Bigg[\sum_{k=1}^{K}\sum_{m=1}^{M}\log\Gamma\left(\sum_{v=1}^{V^{(m)}}\beta_{v}^{(m)}\right)-\sum_{v=1}^{V^{(m)}}\log\Gamma\left(\beta_{v}^{(m)}\right)\\
 & \quad+\sum_{v=1}^{V^{(m)}}\log\Gamma\left(\beta_{v}^{(m)}+n_{\bullet vk}^{(m)}\right)-\log\Gamma\left(\sum_{v=1}^{V^{(m)}}\beta_{v}^{(m)}+n_{\bullet vk}^{(m)}\right)\Bigg]\\
 & \quad+\mathbb{E}_{q\left(z_{(j,-i)}^{(m)}\right)}\Bigg[\sum_{j=1}^{P}\log\Gamma\left(\sum_{k}\alpha_{k}\boldsymbol{\uppi}_{j}\right)-\sum_{k=1}^{K}\log\Gamma\left(\alpha_{k}\boldsymbol{\uppi}_{j}\right)\\
 & \quad+\sum_{k=1}^{K}\log\Gamma\left(\alpha_{k}\boldsymbol{\uppi}_{j}+n_{j\bullet k}^{(\bullet)}\right)-\log\Gamma\left(\sum_{k=1}^{K}\alpha_{k}\boldsymbol{\uppi}_{j}+n_{j\bullet k}^{(\bullet)}\right)\Bigg]\\
 & =\mathbb{E}_{q\left(z_{(j,-i)}^{(m)}\right)}\Bigg[\sum_{v=1}^{V^{(m)}}\log\Gamma\left(\beta_{v}^{(m)}+n_{\bullet vk}^{(m)}\right)-\log\Gamma\left(\sum_{v=1}^{V^{(m)}}\beta_{v}^{(m)}+n_{\bullet vk}^{(m)}\right)\\
 & \quad+\sum_{k=1}^{K}\log\Gamma\left(\alpha_{k}\boldsymbol{\uppi}_{j}+n_{j\bullet k}^{(\bullet)}\right)-\log\Gamma\left(\sum_{k=1}^{K}\alpha_{k}\boldsymbol{\uppi}_{j}+n_{j\bullet k}^{(\bullet)}\right)\Bigg]+\text{const}\\
 & =\mathbb{E}_{q\left(z_{(j,-i)}^{(m)}\right)}\Bigg[\log\Gamma\left(\beta_{x_{ji}^{(m)}}^{(m)}+n_{\bullet x_{ji}^{(m)}k}^{(m)}\right)-\log\Gamma\left(\sum_{v=1}^{V^{(m)}}\beta_{v}^{(m)}+n_{\bullet vk}^{(m)}\right)\\
 & \quad+\log\Gamma\left(\alpha_{k}\boldsymbol{\uppi}_{j}+n_{j\bullet k}^{(\bullet)}\right)-\log\Gamma\left(\sum_{k=1}^{K}\alpha_{k}\boldsymbol{\uppi}_{j}+n_{j\bullet k}^{(\bullet)}\right)\Bigg]+\text{const}\\
 & \overset{(i)}{=}\log\left(\beta_{x_{ji}^{(m)}}^{(m)}+\left[n_{\bullet x_{ji}^{(m)}k}^{(m)}\right]_{(-j,-i)}\right)-\log\left(\sum_{v=1}^{V^{(m)}}\beta_{v}^{(m)}+\left[n_{\bullet vk}^{(m)}\right]_{(-j,-i)}\right)\\
 & \quad+\log\left(\alpha_{k}\boldsymbol{\uppi}_{j}+\left[n_{j\bullet k}^{(\bullet)}\right]_{(j,-i)}\right)-\log\left((\sum_{k=1}^{K}\alpha_{k})\pi_{j}+\sum_{k=1}^{K}\left[n_{j\bullet k}^{(\bullet)}\right]_{(j,-i)}\right)\\
 & \quad+\text{const}\\
 & =\log\left(\left(\alpha_{k}\boldsymbol{\uppi}_{j}+\left[n_{j\bullet k}^{(\bullet)}\right]_{(j,-i)}\right)\frac{\left(\beta_{x_{ji}^{(m)}}^{(m)}+\left[n_{\bullet x_{ji}^{(m)}k}^{(m)}\right]_{(-j,-i)}\right)}{\sum_{v=1}^{V^{(m)}}\beta_{v}^{(m)}+\left[n_{\bullet vk}^{(m)}\right]_{(-j,-i)}}\right)+\text{const}
\end{align*}
The equation $(i)$ follows by defining the first term as
$$\left[n_{\bullet x_{ji}^{(m)} k}^{(m)}\right]_{(-j,-i)}=\sum_{j'=1}^{P}\sum_{i'=1}^{N_{j}^{(m)}}\mathbb{I}\left[(x_{j'i'}^{(m)}=x_{ji}^{(m)},z_{j'i'}^{(m)}=k) \cap (j' \neq j, i' \neq i)\right],$$the second term as
$$\left[n_{\bullet v k}^{(m)}\right]_{(-j,-i)}=\sum_{j'=1}^{P}\sum_{i'=1}^{N_{j}^{(m)}}\mathbb{I}\left[(x_{j'i'}^{(m)}=v,z_{j'i'}^{(m)}=k) \cap (j' \neq j, i' \neq i)\right],$$the third and the forth term as
$$\left[n_{j \bullet k}^{(\bullet)}\right]_{(j,-i)} = \sum_{m=1}^{M}\sum_{i'=1}^{N_{j}^{(m)}}\mathbb{I}\left[(z_{ji'}^{(m)}=k) \cap (i' \neq i)\right].$$

Finally we will get the estimation of the closed-form latent variational expectation update of $\gamma_{jik}^{\left(m\right)}$
after calculating the following and normalizing afterwards:
\begin{align*}
\gamma_{jik}^{\left(m\right)} & \propto\exp\left(\left(\delta_{j}\mf{w}^{\top}\bar{\boldsymbol{\gamma}}_{j}^{\left(m\right)}\right)\left(\delta_{j}\frac{w_{k}}{N_{j}^{\left(m\right)}}\right)\right)\\
 & \quad\times\exp\left[-H_{0}\left(T_{j}\right)\left(\mf{w}^{\top}\bar{\boldsymbol{\gamma}}_{j}^{\left(m\right)}+1\right)\exp\left(\frac{w_{k}}{N_{j}^{\left(m\right)}}\right)\right]\\
 & \quad\times\left(\alpha_{k}\boldsymbol{\uppi}_{j}+\left[n_{j\bullet k}^{(\bullet)}\right]_{(j,-i)}\right)\frac{\left(\beta_{x_{ji}^{(m)}}^{(m)}+\left[n_{\bullet x_{ji}^{(m)}k}^{(m)}\right]_{(-j,-i)}\right)}{\sum_{v=1}^{V^{(m)}}\beta_{v}^{(m)}+\left[n_{\bullet vk}^{(m)}\right]_{(-j,-i)}}
\end{align*}

Furthermore, we update the hyperparameters $\boldsymbol{\upalpha}$ and $\mathcal{B}$ by maximizing the marginal log likelihood function under the estimate of the expectation of the variational parameter. Noting that $\boldsymbol{\upalpha}$ and $\mathcal{B}$ only participate in the unsupervised term of the ELBO, the closed-form update can be derived by the fixed point process \cite{minka2000estimating}:
\begin{align}\label{eq:alpha}
    \alpha_k^* &= \underset{\alpha_k}{\arg \max}  \mathbb{E}_{q(\mathcal{Z})}\left[ p\left(\mathcal{X},\mathcal{Z} \mid \boldsymbol{\alpha},\boldsymbol{\uppi},\mathcal{B}\right) \right] \\
    &= \frac{a_\alpha-1+\alpha_k \sum_{j=1}^P \Psi\left(\alpha_k+n_{j \bullet k}^{\left(\bullet\right)}\right)-\Psi\left(\alpha_k\right)}{b_\alpha+\sum_{j=1}^P \Psi\left(\sum_{k=1}^K \alpha_k+n_{j \bullet k}^{\left(\bullet\right)}\right)-\Psi\left(\sum_{k=1}^K \alpha_k\right)}
\end{align}
\begin{align}\label{eq:beta}
    \beta_{v}^{\left(m\right)*} &= \underset{\beta_v^{(m)}}{\arg \max}  \mathbb{E}_{q(\mathcal{Z})}\left[ p\left(\mathcal{X},\mathcal{Z} \mid \boldsymbol{\alpha},\boldsymbol{\uppi},\mathcal{B}\right) \right] \\ 
    &= \frac{a_\beta-1+\beta_{v}^{(m)}\left(\sum_{k=1}^K \Psi\left(\beta_{v}^{\left(m\right)}+n_{\bullet vk}^{(m)}\right)\right)-K V^{(m)} \Psi\left(\beta_{v}^{(m)}\right)}{b_\beta+\sum_{k=1}^K \Psi\left(V^{(m)} \beta_{v}^{(m)}+\sum_{v=1}^{V^{(m)}} n_{\bullet vk}^{(m)}\right)-K \Psi\left(V^{(m)} \beta_{v}^{(m)}\right)}
\end{align}

To update the survival-relevant parameters $\boldsymbol{w}$ and $h_0(\cdot)$, we focus on maximizing the components related to these parameters within the ELBO. This maximization is conditioned on the expected values of the latent variables $\mathcal{Z}$:

\begin{align}
\left(\mf{w},h_{0}\left(\cdot\right)\right) & =\underset{\mf{w},h_{0}\left(\cdot\right)}{\arg\max}\mathbb{E}_{q(\mathcal{Z})}p\left(\mf{T},\boldsymbol{\updelta}\mid\mathcal{Z},h_{0}(\cdot),\mf{w}\right)\label{eq:survival_coef}\\
 & =\underset{\mf{w},h_{0}\left(\cdot\right)}{\arg\max}\sum_{j=1}^{P}\Bigg\{\delta_{j}\log h_{0}\left(T_{j}\right)+\delta_{j}\mf{w}^{\top}\mathbb{E}_{q(\mathcal{Z})}\left[\bar{\mf{z}}_{j}\right]\\
 & \quad-H_{0}\left(T_{j}\right)\exp\left(\mf{w}^{\top}\mathbb{E}_{q(\mathcal{Z})}\left[\bar{\mf{z}}_{j}\right]\right)\Bigg\}-\lambda_{2}\|\mf{w}\|_{2}^{2}-\lambda_{1}\|\mf{w}\|_{1}\\
 & =\underset{\mf{w},h_{0}\left(\cdot\right)}{\arg\max}\sum_{j=1}^{P}\Bigg\{\delta_{j}\log h_{0}\left(T_{j}\right)+\delta_{j}\mf{w}^{\top}\bar{\boldsymbol{\upgamma}}_{j}\\
 & \quad-H_{0}\left(T_{j}\right)\exp\left(\mf{w}^{\top}\bar{\boldsymbol{\upgamma}}_{j}\right)\Bigg\}-\lambda_{2}\|\mf{w}\|_{2}^{2}-\lambda_{1}\|\mf{w}\|_{1}
\end{align}

Above formula mirrors the coefficients estimates employed in the Cox proportional hazards regression with elastic net penalization, which combines both L1 and L2 norms for regularization \cite{simon2011regularization} . In this context, $\bar{\boldsymbol{\upgamma}}_{j}$ function as covariates, while $\left[T_j, \delta_j\right]_{j=1}^P$ provide the survival information. The update of $\boldsymbol{w}$ and $h_0(\cdot)$ is facilitated using the scikit-survival \cite{sksurv} Python module, a tool specifically designed for handling such statistical computations in survival analysis. 

The whole collapsed variational Inference algorithm for \model~ is in Algorithm \ref{alg:CVI}.

\begin{algorithm}[H]
\caption{Collapsed Variational Inference for MixEHR-SurG}\label{alg:CVI}
\begin{algorithmic}

\State {\textbf{Initialization:}}

     \textrm{$\alpha_k \sim \operatorname{Gamma}(a,b)$ for $k=1, \ldots, K$}

     \textrm{$\beta_v^{(m)} \sim \operatorname{Gamma}(c,d)$ for $v=1, \ldots, V$ and $m=1, \ldots, M$}
     
     \textrm{$\gamma_{j i k}^{(m)} \sim \operatorname{Unif}(0,1)$ for all $i,j,k,m$}

     \textrm{Normalize $\gamma_{jik}^{(m)}$ to sum to 1 over $k$}

\State
\Repeat{\textbf{Converge}}{

\State{\textrm{E-Step:}}

\For{\textrm{$m = 1,\ldots,M$}}{
    
    \For{\textrm{$j = 1,\ldots,P$}}{
        
        \For{\textrm{$i = 1,\ldots,N_j^{(m)}$}}{
        
            \For{\textrm{$k = 1,\ldots,K$}}{            
                Update $\gamma_{jik}^{(m)}$ using Eq. \eqref{eq:gamma}
            }
        
            \textrm{Normalize $\gamma_{jik}^{(m)}$ to sum to 1 over $k$}
            }
        }
    }
\State{\textrm{M-Step:}}

\For{\textrm{$k = 1,\ldots,K$}}{
    \textrm{Update $\alpha_k$ using Eq. \eqref{eq:alpha}}
    }

\For{\textrm{$m = 1,\ldots,M$}}{
    \For{\textrm{$v = 1,\ldots,V^{(m)}$}}{
      \textrm{Update $\beta_v^{(m)}$ using Eq. \eqref{eq:beta}}
    }
}
\textrm{Estimate $\boldsymbol{w}, h_0(\cdot)$ by Eq. \eqref{eq:survival_coef} using Coxnet with updated $\bar{\gamma}_{j}$ as covariates, and survival data $\left[T_j, \delta_j\right]_{j=1}^P$.}
}

\end{algorithmic}
\end{algorithm}

\section{Evaluating causal phenotypes in simulation study}\label{appendix-eval}
For the quantitative evaluation of \model, we first focused on assessing its capability to identify mortality-related topics. In the simulation section,  we used Receiver Operating Characteristic (ROC) curve, a widely-used metric in machine learning to evaluate the variable selection performance of our models. The ROC curve is the true positive rate TPR=TP/(TP+FN) as a function of the false positive rate FPR=FP/(FP+TN) in variable selection, where TP, FP, FN, TN are true positive, false positive, false negative, and true negative, respectively. In our context, this involves comparing the estimated survival coefficients of the simulation data set with the ground truth coefficients we predefined (i.e., 50 survival-related topics with a coefficient of 6, and all others set to 0). 

\section{Survival analysis}\label{appendix-surv}

From $\mf{w}$ learned by \model, we selected the top 3 and bottom 3 survival-related phenotypes with the largest positive and negative coefficients, respectively. To assess the statistical significance of each coefficient $w_k$, we conducted chi-square tests against the null hypothesis that $w_k=0$ \cite{cox1972regression}. Specifically, we divided patients into two groups based on their topic proportion. For the phenotype with the highest survival coefficient, denoted as $k_{\max }=\underset{k}{\arg \max }\ w_k$, we empirically determined the threshold to be the top 30\% percentile of the topic mixture probabilities such that patients above the percentile were assigned to one group and the rest of the patients were assigned to the other group (\fig~\ref{fig:CHD_rlt2}b and \fig~\ref{fig:Mimic_rlt2}b). We then computed the chi-squared test $p$-values using the \texttt{survival} R package \cite{survival-package} (\fig~\ref{fig:CHD_rlt2}c and \fig~\ref{fig:Mimic_rlt2}c).

\section{Supplementary Figures}\label{appendix-fig}

\begin{figure}[h!]
\centering
\includegraphics[width=0.5\textwidth]{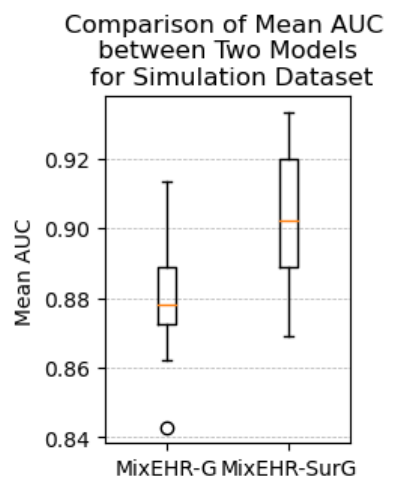}
\caption{Comparison of the mean AUC between the pipeline MixEHR-G+Coxnet and \model~based on 10 simulated datasets.}\label{fig:simbox}
\end{figure}

\begin{figure}[h!]
\centering
\includegraphics[width=1.0\textwidth]{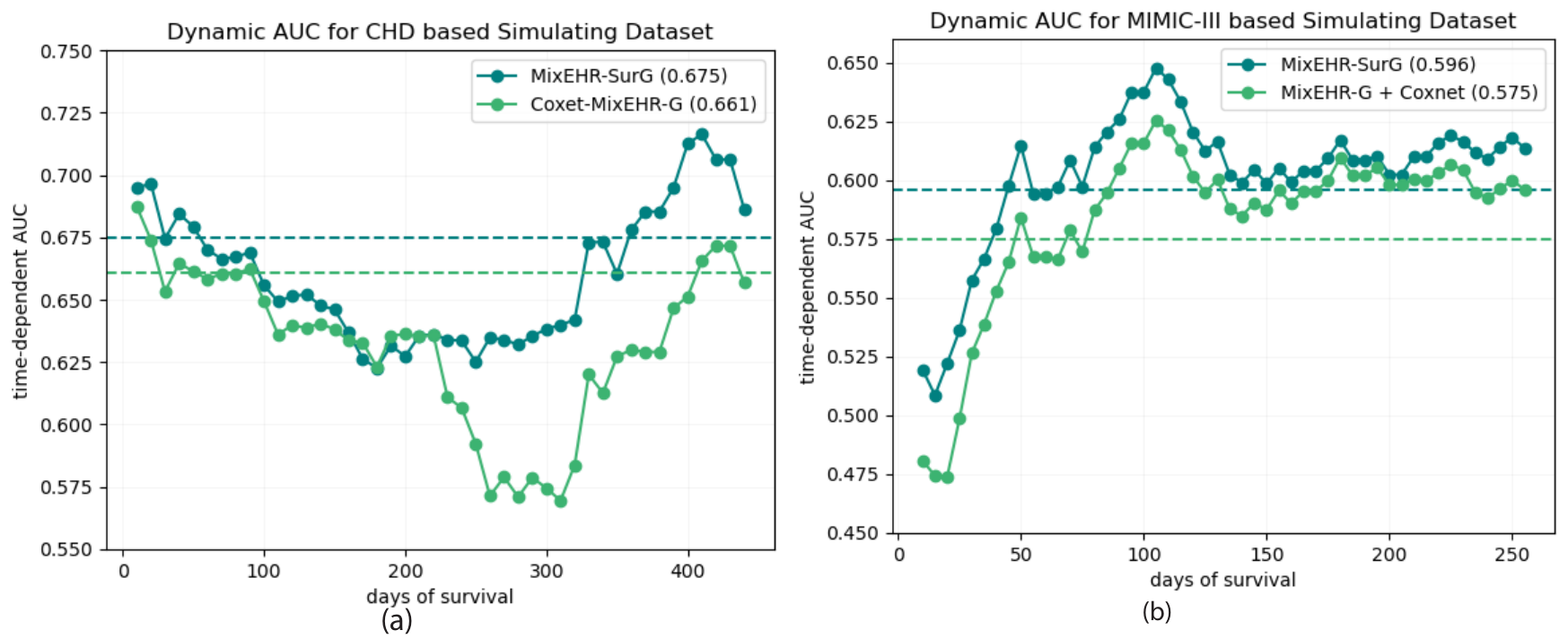}
\caption{Dynamic AUC curves for predicting time to death in patients from the simulated data. (a) Dynamic AUC curves for predicting time to death in patients from simulating dataset based on the CHD dataset. (b) Dynamic AUC curves for predicting time to death in patients from simulating dataset based on the MIMIC-III dataset.}\label{fig:sim2}
\end{figure}

\begin{figure}[h!]
\centering
\includegraphics[width=0.7\textwidth]{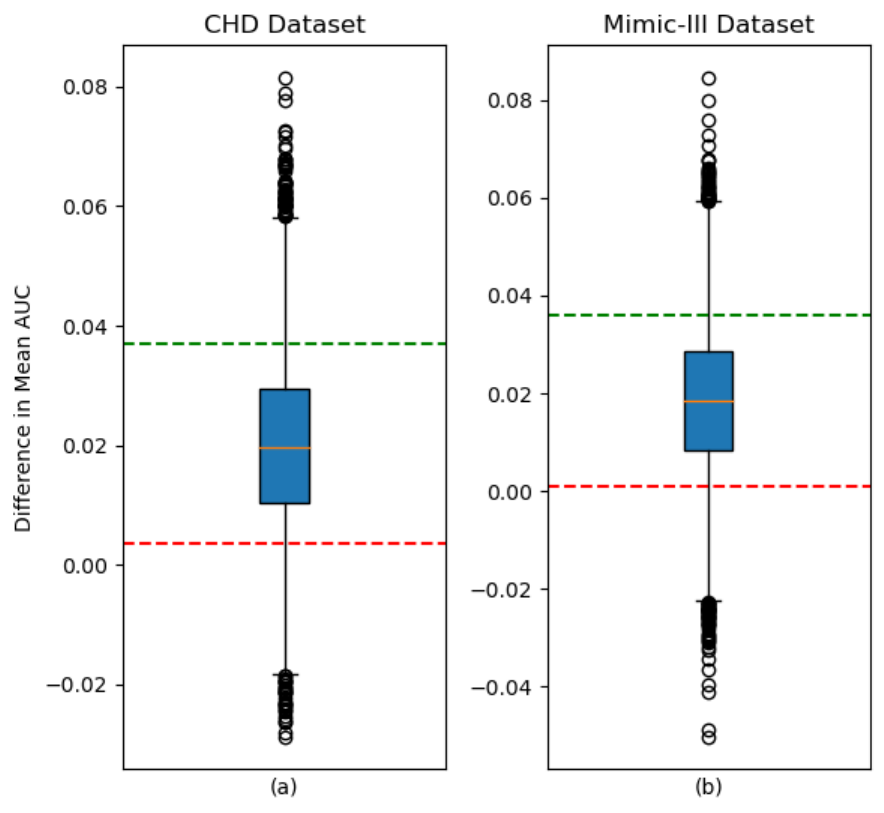}
\caption{\hihi{Comparison of mean AUC differences for mortality time prediction between MixEHR-SurG and MixEHR-G+Coxnet ($\Delta$AUC = AUC(MixEHR-SurG) - AUC(MixEHR-G+Coxnet)), based on 10,000 bootstrap datasets for (a) CHD and (b) MIMIC-III dataset. The 75\% confidence intervals are indicated by the dashed lines.}}\label{fig:bootstrap_boxplot}
\end{figure}

\begin{figure}[h!]
\centering
\includegraphics[width=1.0\textwidth]{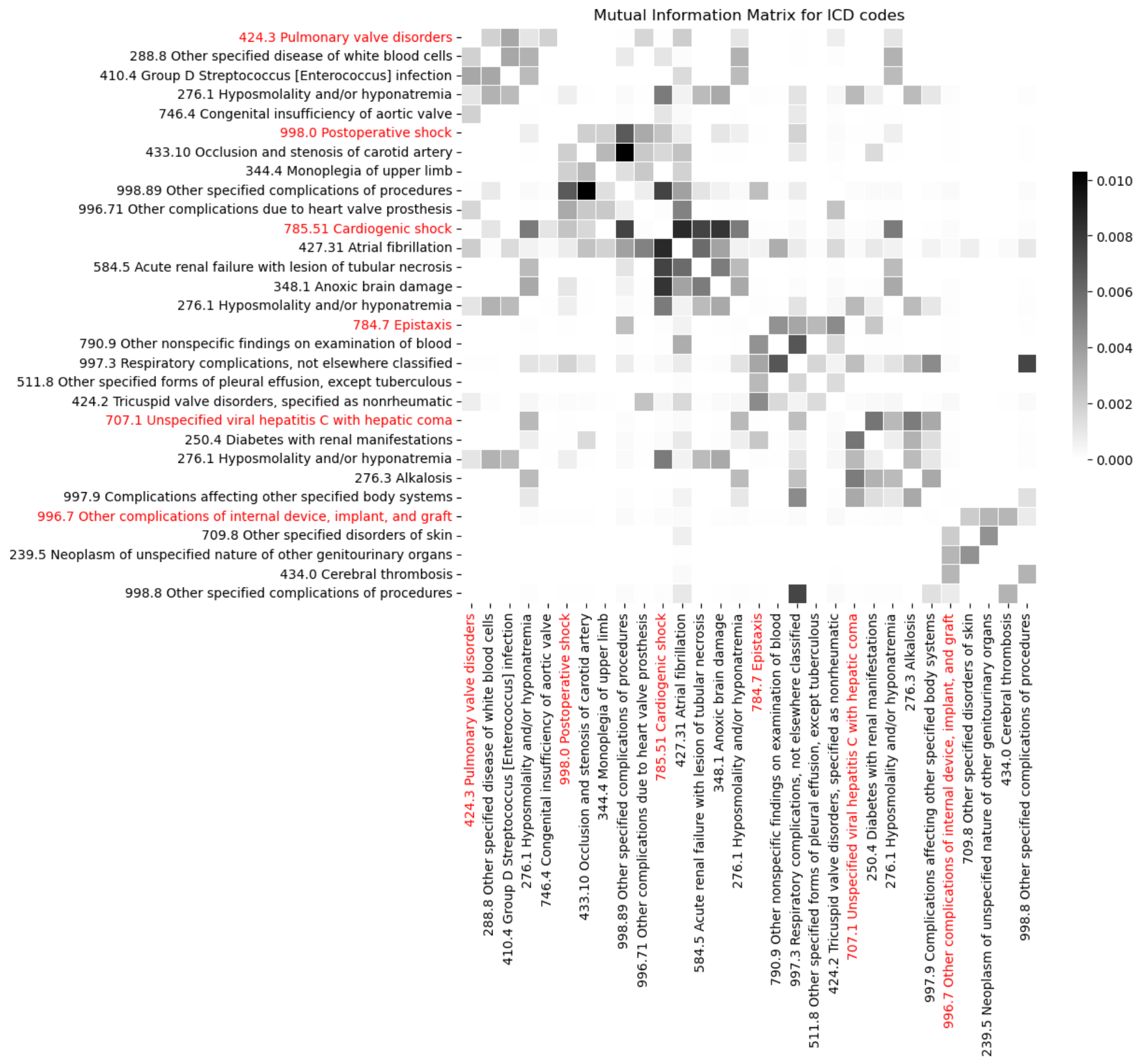}
\caption{Mutual information between the top ICD codes from the top 6 survival phenotype topics identified from the CHD dataset. ICD codes in red are the ones that define the corresponding PheCode. The diagonal entries as well as mutual information between the same ICD codes were intentionally masked out for the ease of viewing.}\label{fig:chd_top_mi}
\end{figure}

\begin{figure}[h!]
\centering
\includegraphics[width=0.9\textwidth]{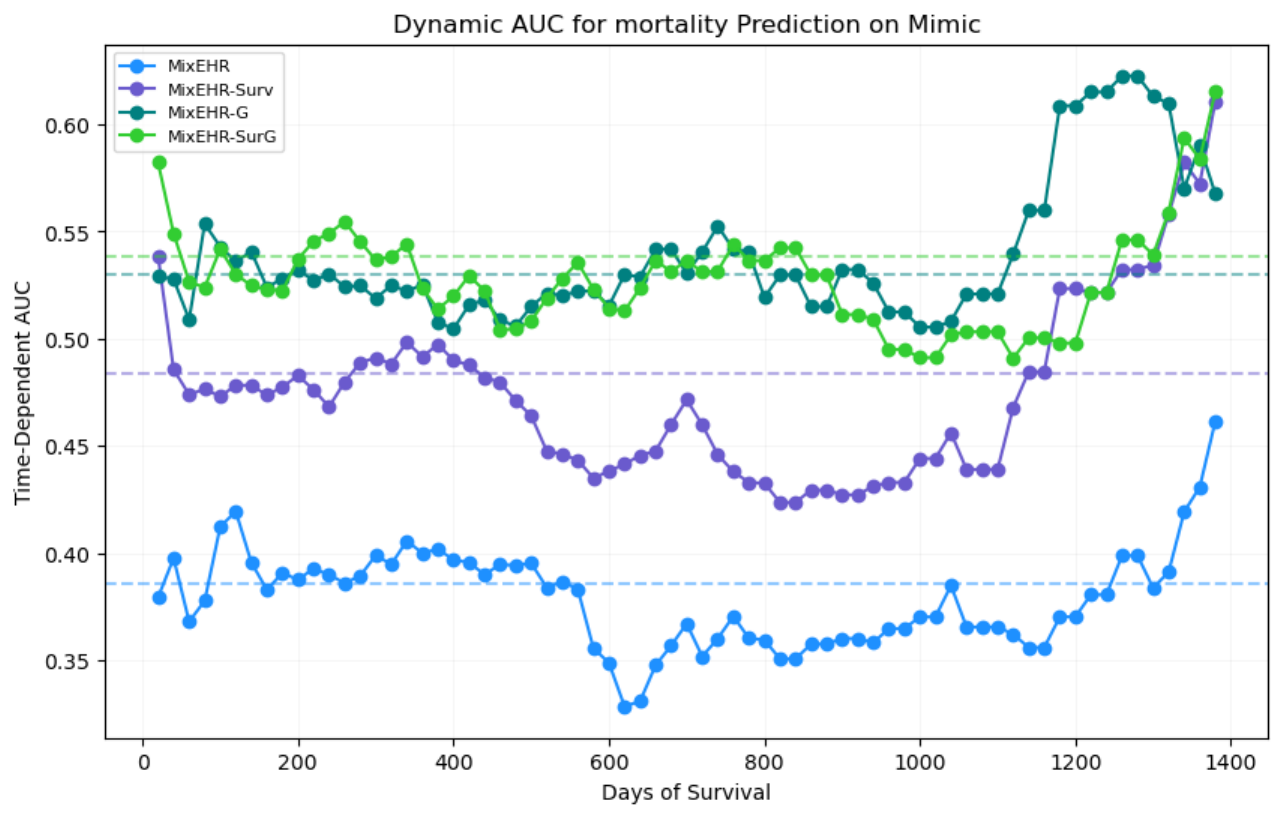}
\caption{Dynamic AUC curves for predicting time to death in patients from the MIMIC-III dataset. We set a series of time points beginning at 20 and increasing in steps of 20, extending to 1400. At each of these intervals, we calculate the cumulative AUC, which is then used to construct the Dynamic AUC curve.}\label{fig:Mimic_rlt1}
\end{figure}

\begin{figure}[h!]
\centering
\includegraphics[width=1.0\textwidth]{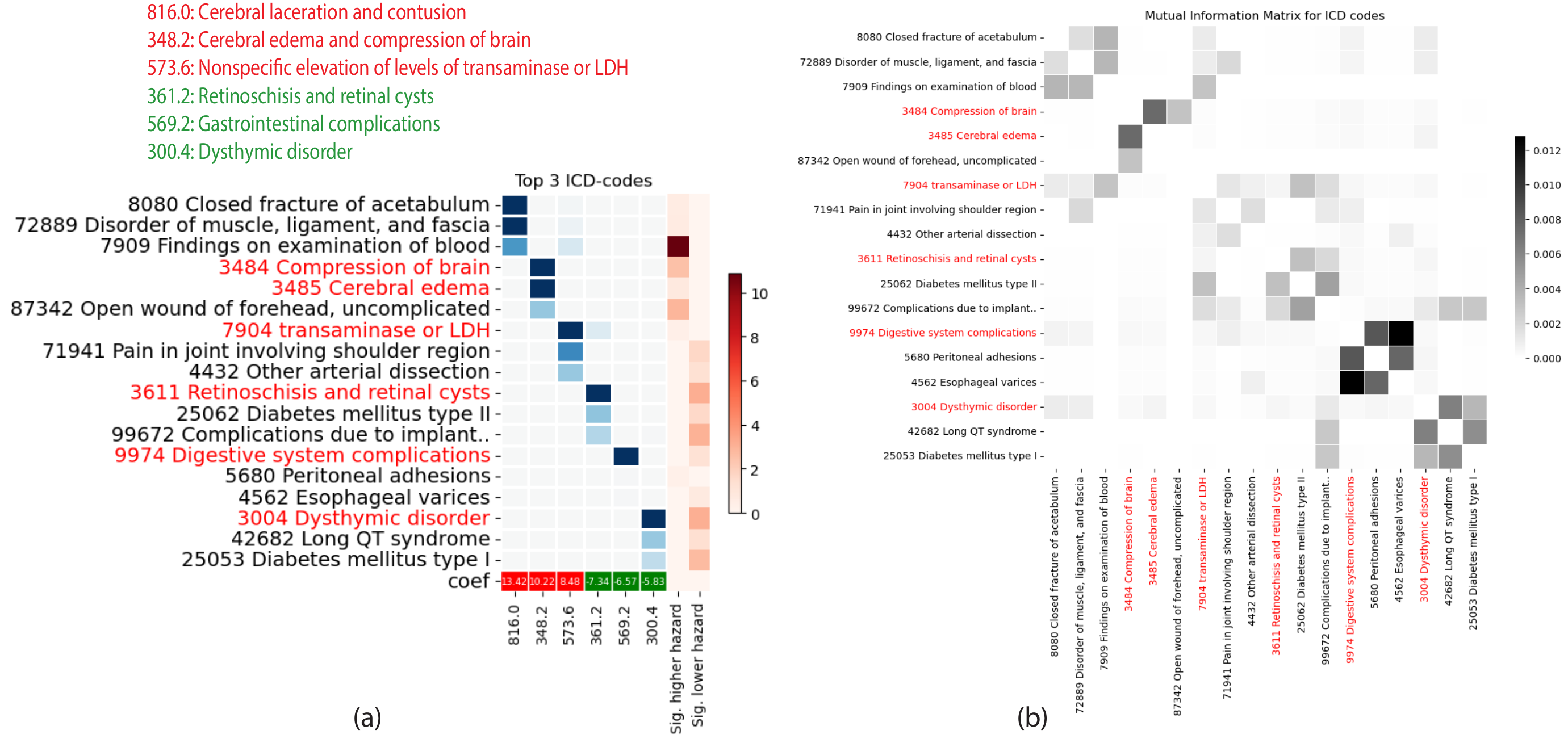}
\caption{Comorbidity analysis of the top ICD codes for survival phenotype topics identified from the MIMIC-III data. (a) Heatmap displaying the top 3 ICD-9 codes per survival phenotype topics for the top 3 and bottom 3 phenotypes. The color gradation indicates the prevalence of each feature within each phenotype topic. The last row indicates the Cox regression coefficients. The last two columns display the color intensities proportional to the -log p-value from the log-rank test for high mortality risk and low mortality risk, respectively. (b) Mutual information between the top ICD codes from the top 6 survival phenotype topics. ICD codes in red are the ones that define the corresponding PheCode. The diagonal entries were intentionally masked out for the ease of viewing.}\label{fig:mimic-icd-heatmap}
\end{figure}

\begin{figure}[h!]
\centering
\includegraphics[width=1.0\textwidth]{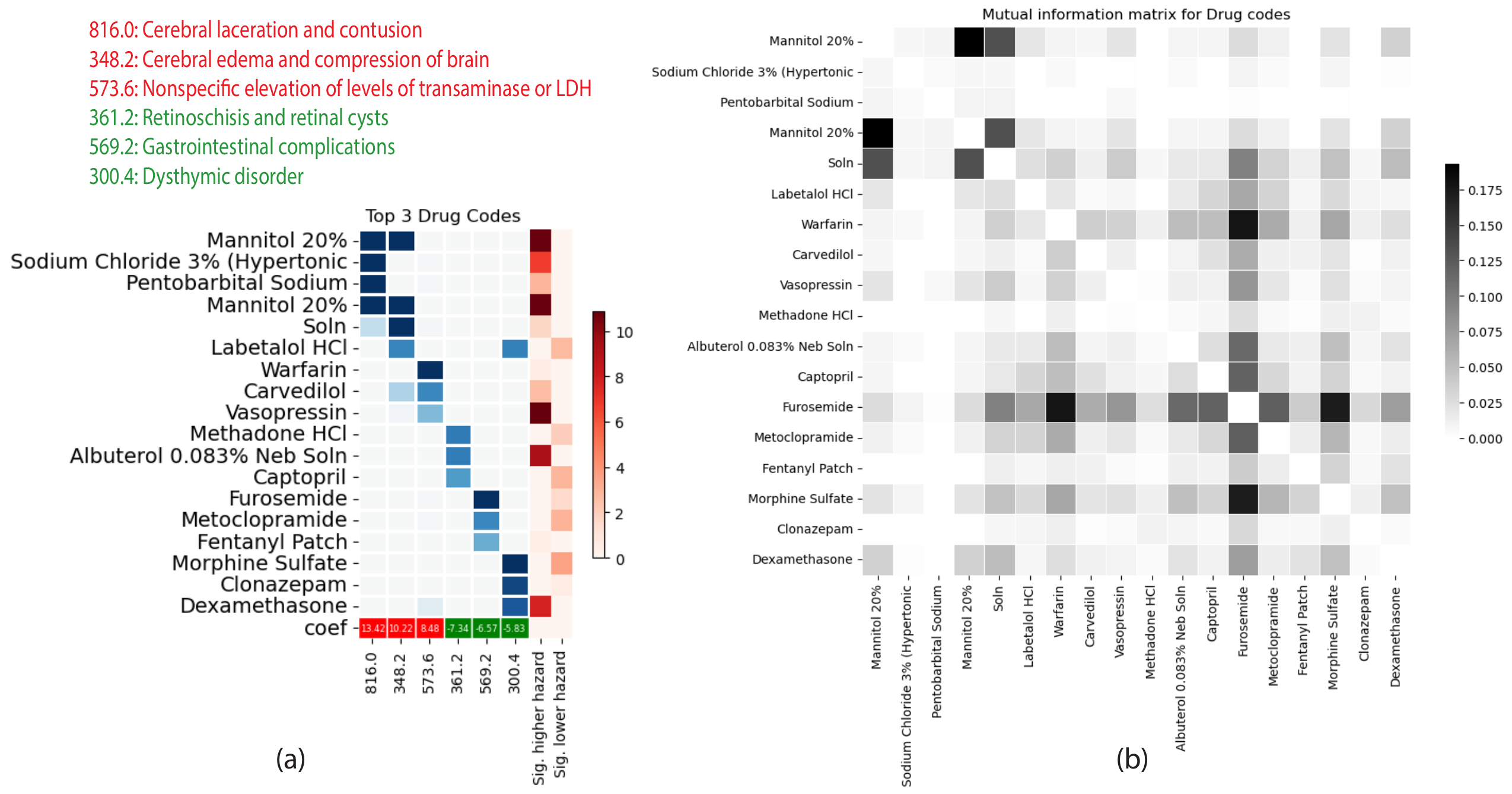}
\caption{Comorbidity analysis of the top drug codes for survival phenotype topics identified from the MIMIC-III data. The presentation of the panels is the same as in \supfig~\ref{fig:mimic-icd-heatmap}}\label{fig:mimic-drug-heatmap}
\end{figure}

\begin{figure}[h!]
\centering
\includegraphics[width=1.0\textwidth]{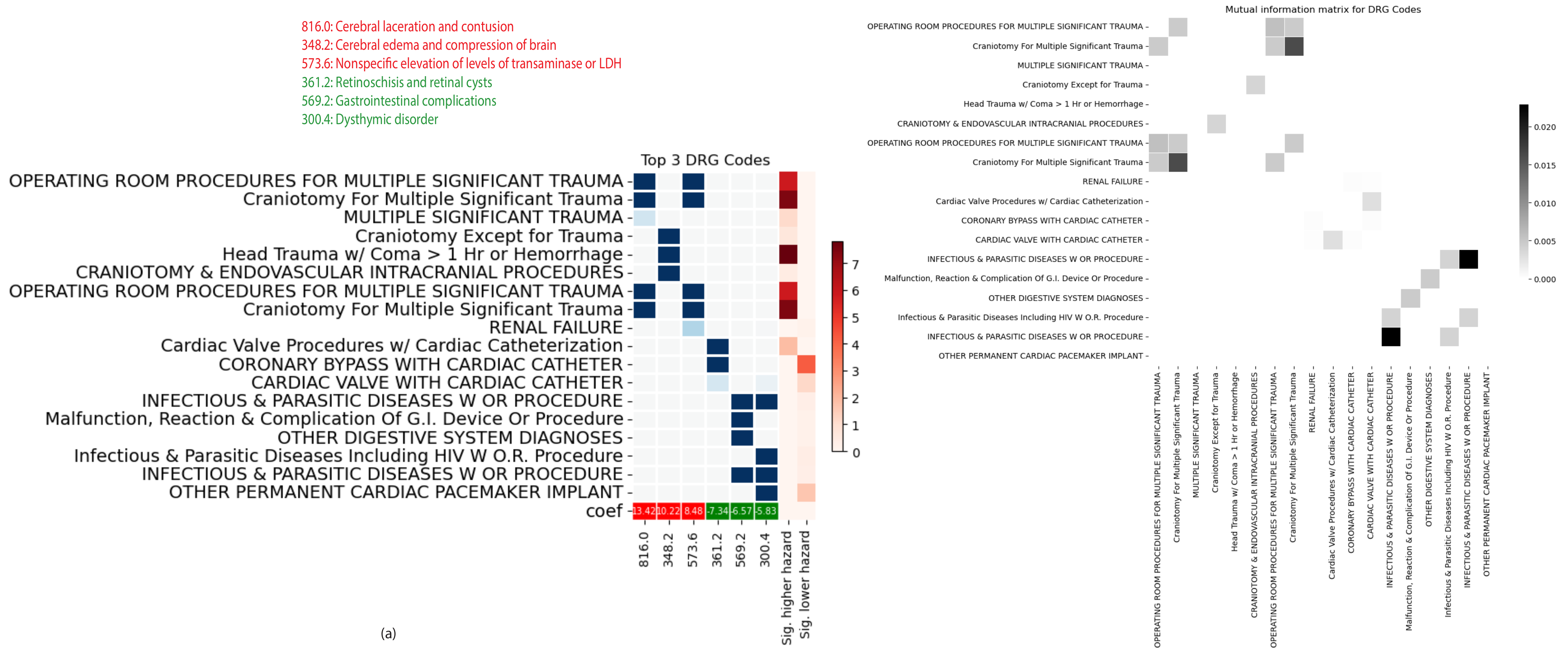}
\caption{Comorbidity analysis of the top DRG codes for survival phenotype topics identified from the MIMIC-III data. The presentation of the panels is the same as in \supfig~\ref{fig:mimic-icd-heatmap}}\label{fig:mimic-DRG-heatmap}
\end{figure}

\begin{figure}[h!]
\centering
\includegraphics[width=1.0\textwidth]{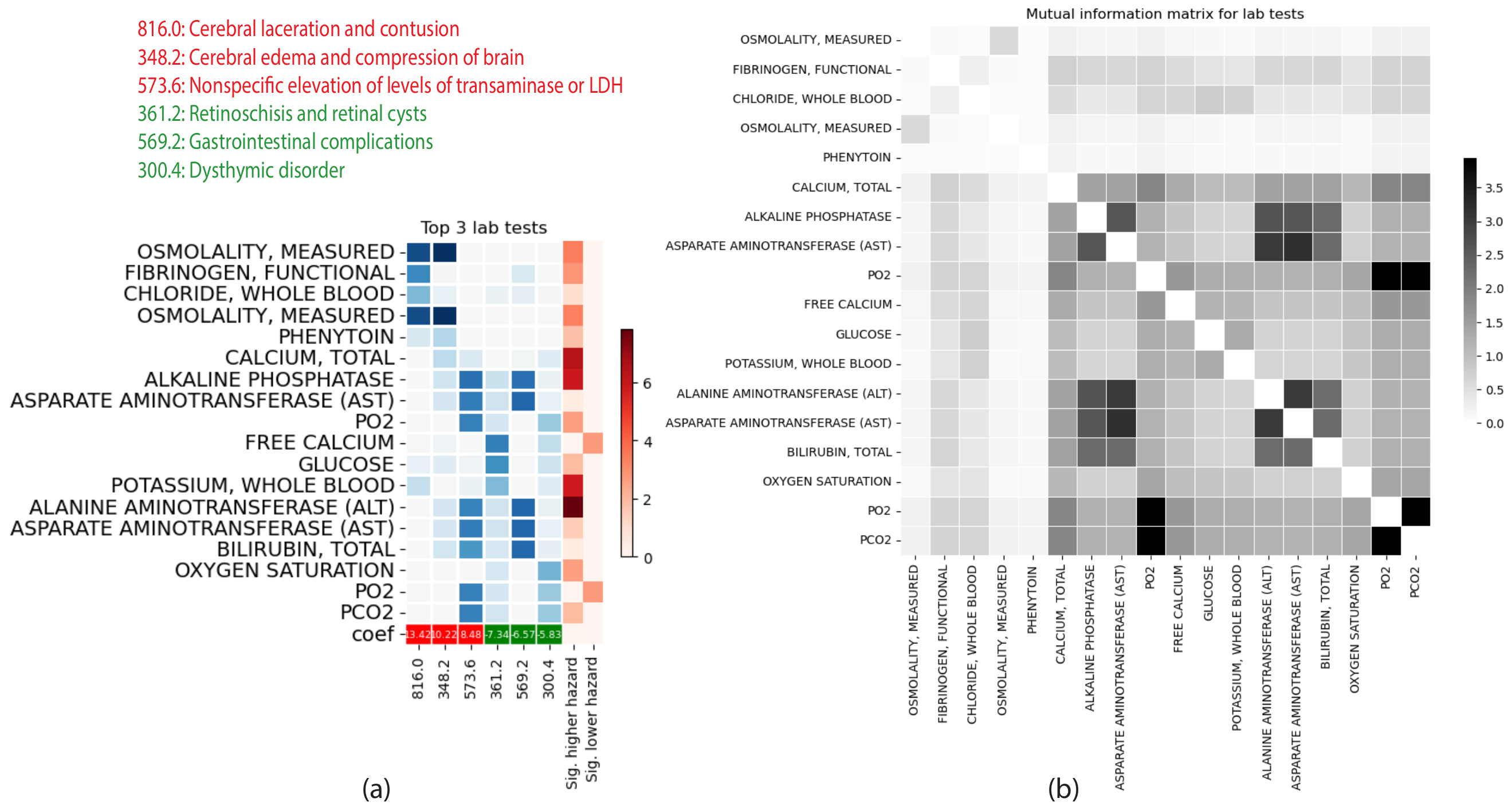}
\caption{Comorbidity analysis of the top lab tests for survival phenotype topics identified from the MIMIC-III data. The presentation of the panels is the same as in \supfig~\ref{fig:mimic-icd-heatmap}}\label{fig:mimic-lab-heatmap}
\end{figure}

\begin{figure}[h!]
\centering
\includegraphics[width=1.0\textwidth]{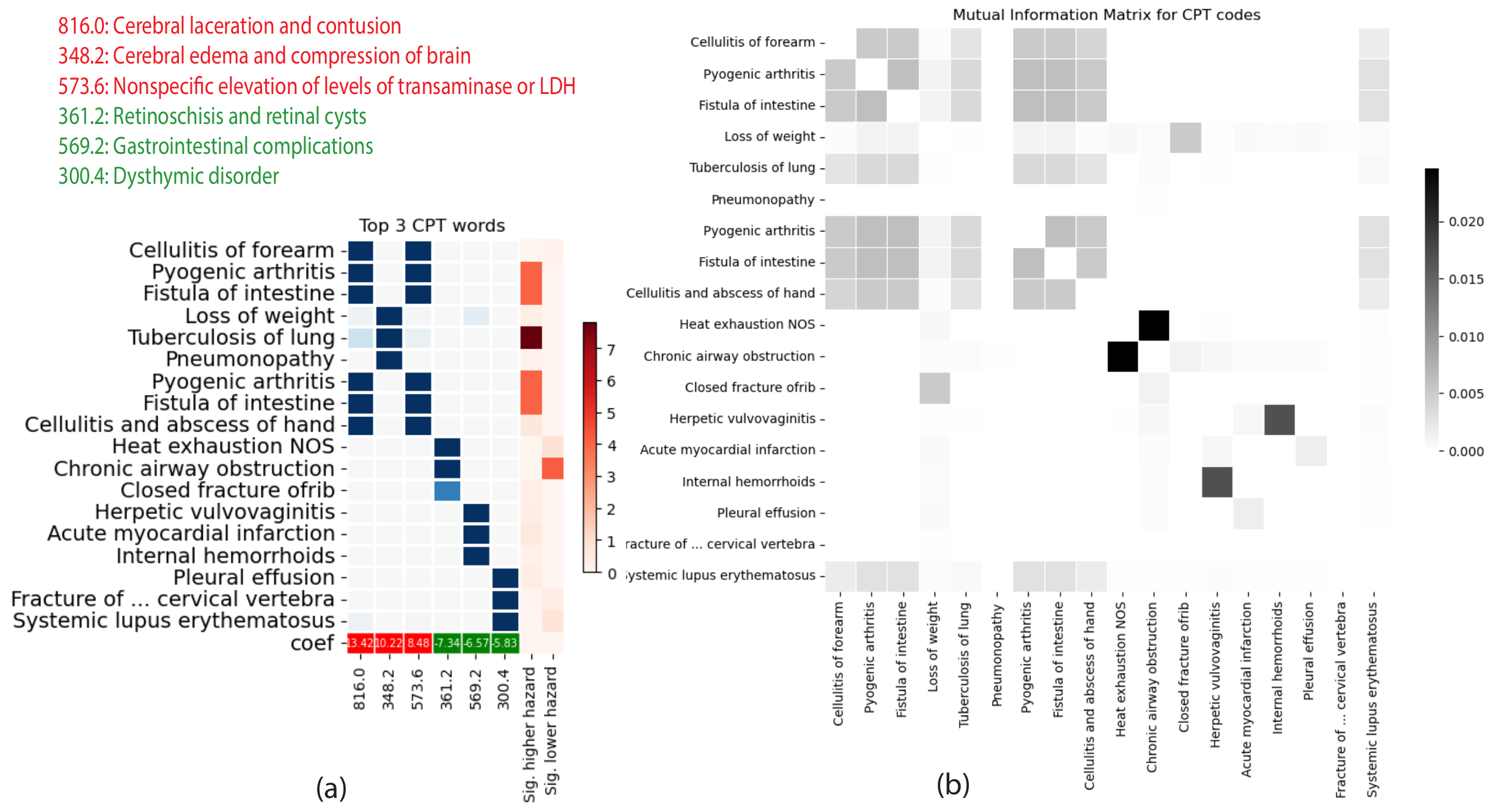}
\caption{Comorbidity analysis of the top CPT words for survival phenotype topics identified from the MIMIC-III data. The presentation of the panels is the same as in \supfig~\ref{fig:mimic-icd-heatmap}}\label{fig:mimic-cpt-heatmap}
\end{figure}

\end{document}